\documentclass{article}
\usepackage{ijcai25}

\usepackage{times}
\usepackage{soul}
\usepackage{url}
\usepackage[hidelinks]{hyperref}
\usepackage[utf8]{inputenc}
\usepackage[small]{caption}
\usepackage{graphicx}
\usepackage{amsmath}
\usepackage{amsthm}
\usepackage{booktabs}
\usepackage{algorithm}
\usepackage{algorithmic}
\usepackage[switch]{lineno}

\usepackage{multirow}
\usepackage{colortbl}
\usepackage{array}
\usepackage{arydshln}

\title{SRA-MCTS: Self-driven Reasoning Augmentation with Monte Carlo \\Tree Search for Code Generation}

\author{Bin Xu\thanks{~~These authors are both First Author.}~~~Yiguan Lin$^{*}$~~~Yinghao Li~~~Yang Gao\thanks{~~Corresponding author} \\
School of Computer Science and Technology, Beijing Institute of Technology, Beijing, China \\
\{binxu,yglin,yhli,gyang\}@bit.edu.cn}

\begin{document}

\maketitle

\begin{abstract}
% 200词摘要
% 大型语言模型在简单的代码生成任务中表现出卓越的性能。然而，在处理需要推理和问题分解的复杂问题时，它们会遇到重大挑战。为了解决这个问题，我们提出了一种自驱动推理增强程序--SRA-MCTS，它结合了蒙特卡洛树搜索（MCTS）来生成推理数据。
% SRA-MCTS 使 LLM 能够自我生成中间推理步骤并执行迭代自我评估，从而促进自我完善。具体来说，它利用 MCTS 生成各种中间推理步骤。在每次迭代过程中，MCTS 生成一个步骤，并利用自我评价来指导后续分支的选择，最终形成一个足够多样化的推理路径，称为 “思考”。这种思考指导模型生成相应的代码，两者结合起来作为训练数据进行监督微调。
% 实验结果表明，SRA-MCTS 在三种规模的模型中都取得了一致的性能改进，无需额外的监督辅助。在应用于 Meta-Llama-3.1-8B-Instruct 模型时，它在 MBPP-Complex 数据集上提高了 11 分，突出了模型自我改进的巨大潜力。

Large language models exhibit remarkable performance in simple code generation tasks. However, they encounter significant challenges when addressing complex problems that require reasoning and question decomposition. To tackle this, we propose a self-driven reasoning augmentation process, SRA-MCTS, which incorporates Monte Carlo Tree Search (MCTS) for reasoning data generation. 
SRA-MCTS enables LLMs to self-generate intermediate reasoning steps and perform iterative self-evaluation, facilitating self-improvement. Specifically, it utilizes MCTS to produce diverse intermediate reasoning steps. During each iteration, MCTS generates a step and employs self-evaluation to guide the selection of subsequent branches, ultimately forming a sufficiently diverse reasoning path referred to as “thinking”. This thinking guides the model in generating corresponding code, and both are combined as training data for supervised fine-tuning. 
Experimental results demonstrate that SRA-MCTS achieves consistent performance improvements across three model scales without additional supervisory assistance. Applied to the Meta-Llama-3.1-8B-Instruct model, it delivers an 11-point improvement on the MBPP-Complex dataset, underscoring the significant potential for model self-improvement.

\end{abstract}

% glossary
% self-generated
% question, thinking, code

\section{Introduction}

% 1. Introduction

\begin{figure}[h]
    \centering    \includegraphics[width=\linewidth]{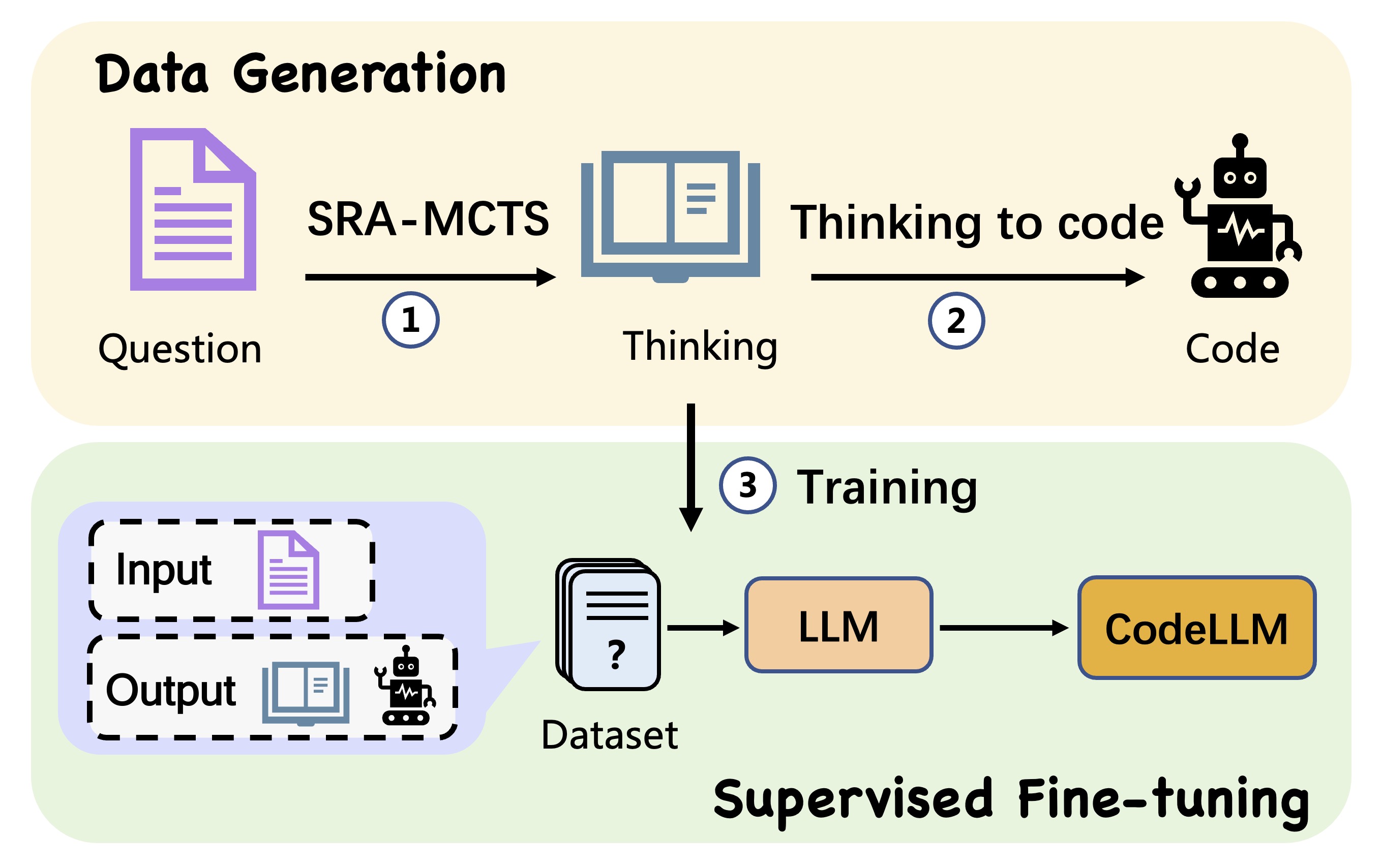}
    \caption{The overall workflow of our method, with data generation shown at the top and training at the bottom. SRA-MCTS guides the LLM to generate thinking, which is then used by the LLM as a part of the prompt to generate the corresponding code. The question, thinking, and code are organized as training data for supervised fine-tuning.
    }
    \label{fig:pipeline}
\end{figure}
% \begin{figure}[h]
%     \centering
%     \includegraphics[width=1\linewidth]{intro.png}
%     \caption{The figure compares the impact of different reasoning augmentation methods on model performance. ``Instruct'' refers to the officially released model, while ``MCTS-Infer'' represents results obtained by applying MCTS-based reasoning to the official model. ``CoT-Trained'' and ``ToT-Trained'' denote the results of fine-tuning with data generated using CoT-Trained or ToT-Trained methods, respectively, and ``SRA-MCTS-Trained'' represents the results of fine-tuning with data generated using the SRA-MCTS method.}
%     % 图中比较了不同推理增强方法生成的数据对模型性能的影响。Instruct版本为官方发布的版本，CoT和ToT为我们使用CoT或ToT生成数据微调的版本，SRA-MCTS为我们使用SRA-MCTS数据生成方法生成数据微调的版本。
%     \label{fig:intro}
% \end{figure}

% Remake Version
% 大问题
% 现在的大模型在代码生成领域已经能够很好地解决简单和中等的问题，但在复杂问题上表现不佳。
% 小问题的工作
% 有些工作通过GPT来增强数据中的问题部分，通过对开源数据集进行深化提升或者对开源代码进行总结归纳来增强问题的深度和广度。
% 小问题
% 但是，只对数据中问题的改进或提升，能得到高质量的question，没有对回答进行有效提升，不能保证回答的正确性。
Large language models (LLMs) excel at generating code for simple questions, but their performance on more complex tasks remains unsatisfactory.~\cite{DBLP:conf/iclr/LuoX0SGHT0LJ24,DBLP:journals/corr/abs-2305-10679}. 
% Some approaches improve the question portion of datasets by leveraging GPT-3.5 to refine existing datasets or summarize open-source code, thereby enhancing the depth and breadth of the questions~\cite{DBLP:conf/nips/Ouyang0JAWMZASR22,DBLP:conf/iclr/LuoX0SGHT0LJ24,DBLP:journals/corr/abs-2312-02120}. However, improving only the questions in the data leads to high-quality questions, but without special treatment for the answers, the improvement remains limited.
% 最近的研究通常通过提升训练集的质量来训练性能更高的模型，而对训练数据的提升通常包含两个方面。
Recent studies have predominantly concentrated on enhancing the quality of training datasets to improve model performance. These improvements typically address two fundamental aspects of the training data.

% 一方面，通过提升数据的问题质量来加深模型对输入的理解程度。一些方法通过利用性能更高的模型或对已有的开源代码进行总结，来优化数据集的问题部分，从而增强问题的深度和广度。
On one hand, improving the quality of the questions enhances the model's understanding of the input.
Some methods optimize the question portion of the dataset by leveraging more powerful models or summarizing existing open-source code, thereby enhancing both the depth and breadth of the questions~\cite{DBLP:conf/nips/Ouyang0JAWMZASR22,DBLP:conf/iclr/LuoX0SGHT0LJ24,DBLP:journals/corr/abs-2312-02120}.
% 另一方面，通过提升数据答案的复杂程度来引导模型输出质量更高的代码。BRAINSTORM引入了头脑风暴来生成并选择有关问题的不同想法，以增强模型推理。DolphCoder利用GPT-3.5使用多个prompt template得到多样化的输出来进行多目标指令微调。这些方法都旨在使用一些额外的自然语言文本来引导LLM产生高质量的结果。
On the other hand, increasing the complexity of the answers helps guide the model to generate higher-quality codes. 
BRAINSTORM~\cite{DBLP:journals/corr/abs-2305-10679} introduces a brainstorming approach to generate and select diverse ideas related to a given question, thereby enhancing the model's reasoning capabilities. DolphCoder~\cite{DBLP:conf/acl/Wang0DWZDXWZC24} utilizes GPT-3.5 with multiple prompt templates to obtain diverse outputs for multi-objective instruction fine-tuning. These methods all aim to leverage additional natural language text to guide LLMs in generating high-quality results.

% 回答相关的工作
% 有一些工作通过引入自然语言中间过程作为回答的一部分来增强数据中的回答部分，引导模型向正确的方向生成。有些使用GPT以CoT形式对问题分几次生成自然语言解决方案，有些使用简单的idea来引导。希望能通过自然语言这一中间形式引导大模型的思考方向来生成正确结果。
% Some studies direct the model to generate results in the correct direction, improving the answers in the data by incorporating a natural language intermediary process as part of the response. DolphCoder~\cite{DBLP:conf/acl/Wang0DWZDXWZC24} utilize GPT-3.5 to generate natural language solutions in a step-by-step Chain-of-Thought (CoT) format. BRAINSTORM~\cite{DBLP:journals/corr/abs-2305-10679} introduce guiding ideas to steer the process. 
% \textcolor{blue}{Previous studies aim to guide the reasoning process of LLMs toward generating high-quality results by using natural language as an intermediary. }

% 已经有一些工作沿着这个思路进行并取得了一些进展。ScaleAI验证了向大模型以自然语言形式提供正确的解决方案能够大幅提升模型的性能，即使这些自然语言是只有数十个token的不完整的方案。这证明了自然语言解决方案能够引导并启发大模型向正确的方向思考。然而，这些工作缺乏对这些自然语言方案的解释，并且没有内在的纠正错误的机制，从而限制了模型对于自然语言推理步骤的利用。因此我们从模型生成的中间自然语言推理步骤入手，为模型提供一个可靠的具有足够多样性的推理方向。 
Furthermore, ScaleAI~\cite{DBLP:journals/corr/abs-2409-03733} has demonstrated that providing LLMs with correct solutions in natural language form can significantly enhance model performance, even when these solutions consist of incomplete plans with only a few dozen tokens. This highlights the potential of natural language solutions to guide and inspire LLMs to think in the right direction.
Similar to ScaleAI, there have been some works exploring how to improve the intermediate reasoning steps to improve model performance.~\cite{DBLP:conf/cvpr/0003XKISGX24,DBLP:journals/corr/abs-2409-09584,DBLP:conf/acl/LongWXZDCW24}.
However, these studies lack explanations for the natural language solutions and lack an inherent mechanism for error correction, limiting the model's ability to effectively utilize natural language reasoning steps. Therefore, we focus on the intermediate natural language reasoning steps generated by the model, providing it with a reliable and diverse reasoning direction.

In this paper, we refer to the intermediate natural language reasoning steps generated by the model for code generation as ``thinking''. We propose a Self-driven Reasoning Augmentation method, leveraging Monte Carlo Tree Search (MCTS)~\cite{DBLP:conf/cg/Coulom06} to generate high-quality thinking, which we call SRA-MCTS as shown in Figure~\ref{fig:pipeline}. 
Different from ReST-MCTS*~\cite{DBLP:journals/corr/abs-2406-03816}, which uses MCTS to generate complete answers, we apply MCTS to the more critical thinking portion. This provides the model with reliable reasoning, enhancing its ability to generate high-quality code.
SRA-MCTS iteratively constructs the final reasoning path for a given question. Specifically, each iteration generates multiple potential next steps based on the current reasoning path. During each step selection, the model reflects on and scores the new generated step, updating the scores of previously generated steps. This process allows the model to re-evaluate and reconsider unselected steps in subsequent iterations.
This iterative process enables the model to engage in self-reflection, correcting potentially flawed reasoning steps. Finally, the sequence of reasoning steps generated over multiple iterations is combined into comprehensive thinking that guides the model in generating the final code. 
The data generated through SRA-MCTS will be used to fine-tune the model, ultimately resulting in a higher-performing code generation model.
Our contributions are as follows:

\begin{itemize}
    \item We propose a plug-and-play reasoning augmentation data generation method, SRA-MCTS. This method is simple and effective, as it enables the LLM to self-generate and self-evaluate data during the process. Then the generated data is used for model training, fostering the model’s self-improvement.
    % 我们针对code领域提出了一个pipeline，问题由SRA-MCTS生成多样自然语言plan。
    % 该方法由大模型自主完成，无需额外监督，引导模型自我生成高质量的中间推理路径.形成正反馈循环，实现持续提升。
    \item We propose a pipeline for the code generation domain, where input questions are processed by SRA-MCTS to generate diverse, high-quality thinking without additional supervision, thereby creating a positive feedback loop for continuous improvement.
    % 我们对生成的数据进行了详细分析，分析了数据的质量和长度如何影响模型性能。并且通过模型在原benchmark和划分出的更复杂的benchmark上的评测表现对比，验证了推理增强这一方法的有效性。
    \item We perform a detailed analysis of the generated data, examining the impact of its quality on model performance. Additionally, by comparing the model’s performance on the original benchmark and the derived, more complex benchmarks, we validate the effectiveness of the self reasoning augmentation approach.
\end{itemize}

\section{Related Work}
% 代码生成领域中的数据蒸馏
% 数据蒸馏是一种获得高质量数据的有效方法，在代码领域中通常针对问题和答案两个角度进行蒸馏。对问题的蒸馏中，CodeAlpaca使用self-instuct通过将prompt和seed task修改成代码相关，从GPT获得了20K条问题扩展的指令遵循数据。Evol-Instruct加入了各种启发式规则，如推理难度，时空复杂度，在CodeAlpaca的基础上使用GPT增加种子指令的复杂性和多样性，成功增加了数据的深度和广度。OSS-Instruct利用了广泛可获取的开源代码片段，通过设计prompt让GPT从所给片段中的若干行提炼出对应的问题，并且在深度和广度上进行深入，使问题更符合真实世界的分布。CodeOcean使用基于LLM的生成器和判别器，同样对开源代码片段使用GPT3.5进行蒸馏，使用GPT4来进行判别，得到了一批高质量且多样的数据。

% 数据增强
% Evol-Instruct加入了各种启发式规则，如推理难度，时空复杂度，在CodeAlpaca的基础上使用GPT增加种子指令的复杂性和多样性，成功增加了数据的深度和广度。OSS-Instruct利用了广泛可获取的开源代码片段，通过设计prompt让GPT从所给片段中的若干行提炼出对应的问题，并且在深度和广度上进行深入，使问题更符合真实世界的分布。CodeOcean使用基于LLM的生成器和判别器，同样对开源代码片段使用GPT3.5进行蒸馏，使用GPT4来进行判别，得到了一批高质量且多样的数据。DolphCoder基于给定的问题，使用GPT对问题进行分步CoT的自然语言解答，降低了大模型对数据学习的难度。

\paragraph{Data Augmentation} Evol-Instruct~\cite{DBLP:conf/iclr/LuoX0SGHT0LJ24} enhances CodeAlpaca~\cite{codealpaca} by integrating heuristic rules, which increases the complexity and diversity of the seed instructions. OSS-Instruct~\cite{DBLP:journals/corr/abs-2312-02120} leverages open-source code snippets to generate questions that better represent real-world distributions. CodeOcean~\cite{DBLP:conf/acl/YuZSHXZHY24} utilizes GPT-3.5 and GPT-4 for question distillation and filtering, resulting in high-quality, diverse data. DolphCoder~\cite{DBLP:conf/acl/Wang0DWZDXWZC24} improves LLM learning by generating step-by-step answers through multiple turns and pseudo code for problem-solving.
% Evol-Instruct~\cite{DBLP:conf/iclr/LuoX0SGHT0LJ24} improves CodeAlpaca~\cite{codealpaca} by incorporating heuristic rules, thereby increasing the complexity and diversity of seed instructions. OSS-Instruct~\cite{DBLP:journals/corr/abs-2312-02120} utilizes open-source code snippets to generate questions that more accurately reflect real-world distributions. CodeOcean~\cite{DBLP:conf/acl/YuZSHXZHY24} employs GPT-3.5 and GPT-4 for question distillation and filtering, producing high-quality, diverse data. DolphCoder~\cite{DBLP:conf/acl/Wang0DWZDXWZC24} enhances LLM learning by generating step-by-step CoT answers through multiple turns and pseudo code for problems.

% 内外数据合成
% 大模型自我合成数据，能有效缓解语言风格带来的理解偏差。ReCo\cite{DBLP:conf/acl/LiZS24}使用大模型重写代码库中的代码以减小检索时因stylistic deviation导致的retrieval accuracy下降。ReST-MCTS∗开发了一种强化自训练方法，基于将过程奖励指导与MCTS∗相结合，以收集更高质量的推理轨迹以及每步值来训练策略和奖励模型。
\paragraph{Self-generated Data} Data generated by LLMs can effectively mitigate biases in understanding arising from variations in language style. ReCo~\cite{DBLP:conf/acl/LiZS24} utilizes LLMs to rewrite code from codebases, thereby reducing retrieval accuracy degradation caused by stylistic deviations during the retrieval process. ReST-MCTS*~\cite{DBLP:journals/corr/abs-2406-03816} introduces a reinforced self-training approach that combines process reward guidance with MCTS to collect higher-quality reasoning trajectories and stepwise values, which are subsequently used to train the strategy and reward models.
% 探索与思考方法
% 类似于人类在回答难题之前可能会思考很长时间，o1 在尝试解决问题时会使用一系列思维。通过强化学习，o1 学会磨练其思维链并完善其使用的策略。它学会认识并纠正错误。它学会将棘手的步骤分解为更简单的步骤。 AFLOW使用蒙特卡罗树搜索进行有效地探索，通过代码修改、树结构体验和执行反馈迭代地完善工作流程。StepCoder利用规范解决方案的一部分作为提示，使 LLM 能够从简单序列开始探索，并使用FeedBack进行强化学习
\paragraph{Exploration and Thinking Methods} Similar to how humans think deeply before answering difficult questions, OpenAI o1~\cite{openai_learning_to_reason} employs a series of thought processes to solve problems. Through reinforcement learning, it refines its strategies, corrects errors, and streamlines complex steps. AFLOW~\cite{DBLP:journals/corr/abs-2410-10762} employs MCTS to refine workflows through iterative code modifications and feedback. StepCoder~\cite{DBLP:conf/acl/Dou0JZXSHWFXZJZ24} utilizes parts of a standardized solution as a prompt, enabling LLMs to explore simple sequences and improve through feedback-based reinforcement learning.

% 我们的SRA-MCTS不仅对数据的thinking process进行了增强，提高了模型的reasoning和thinking能力，并且达成了全流程的模型自主：自我生成，自我评估，
Unlike previous works, our SRA-MCTS not only enhances the thinking process within the data, improving the model’s reasoning and thinking capabilities, but also establishes a fully autonomous pipeline that includes self-generation, self-evaluation, and ultimately self-improvement.

\section{Method}
% 我们提出了一个利用自身模型来增强训练数据从而提升性能的pipeline，包括SRA-MCTS to plan，plan to code，training。我们为code提供了自然语言的描述生成，从而enhance了模型在代码生成领域的性能。
% 我们的方法分为数据生成和训练，如图所示。对于一批问题，先让大模型使用SRA-MCTS生成针对问题的自然语言分步解决方案（除了必要的公式，其中不包括代码），再让大模型针对问题和生成的解决方案来生成具体的代码，将问题，自然语言解决方案，代码拼接起来形成微调数据集，用于大模型的训练。
\begin{figure*}[!h]
    \centering
    \includegraphics[width=0.95\textwidth]{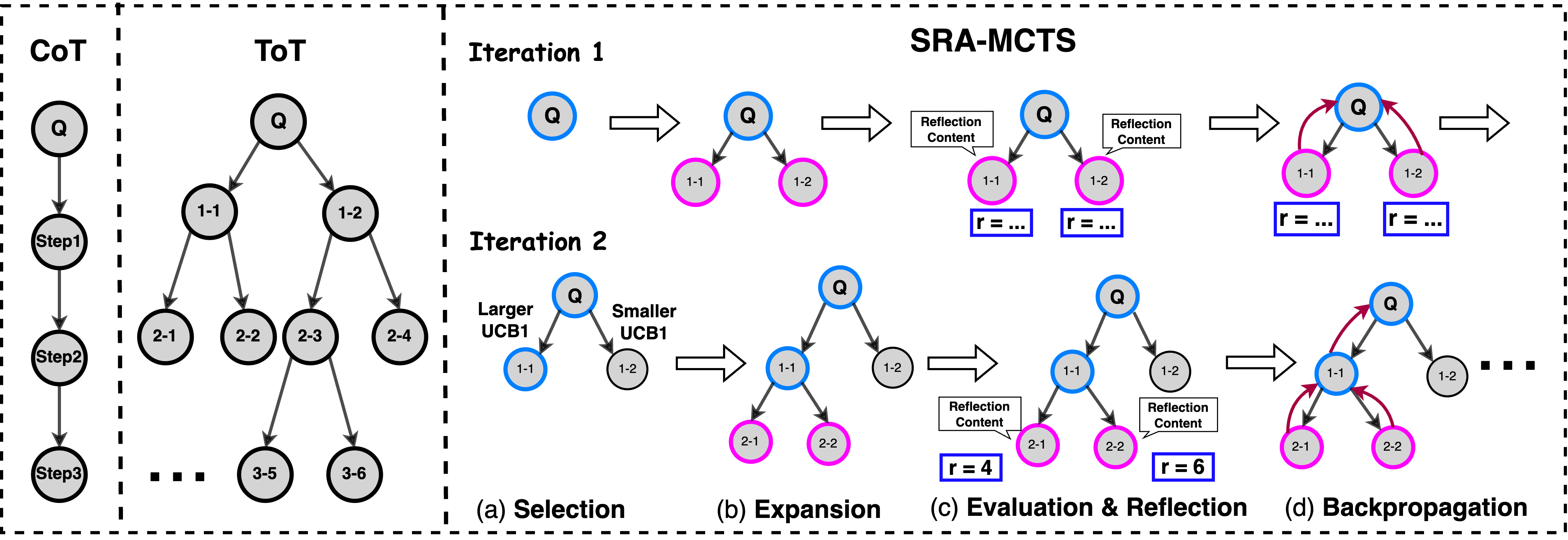}
    \caption{Self-driven reasoning augmentation process of SRA-MCTS. (a) Selection: A leaf node is selected to be expanded in the next phase. (b) Expansion:  A single step is generated and assigned to the node. (c) Evaluation \& Reflection: The step in the node is scored and an insight is generated as reflection. (d) Backpropagation: Reward scores are propagated back. In the notation ``1-1'' within a node, the first ``1'' indicates that it is the 1-st step in the thinking, and the second ``1'' denotes the 1-st variant for this step. The same logic applies to other nodes. \textcolor{blue}{Blue} nodes represent the selected nodes, and \textcolor{red}{red} nodes represent newly expanded nodes.}
    \label{fig:mcts}
\end{figure*}
We propose a pipeline that leverages the model itself to enhance training data and augment reasoning ability, as shown in Figure~\ref{fig:pipeline}, consisting of three stages: SRA-MCTS, Thinking to code, and Training. First, we use the LLM with SRA-MCTS to generate step-by-step natural language thinking for each question. Then, the LLM generates specific codes based on the question and the generated thinking. Finally, we combine the question, the thinking, and the code to create a fine-tuning dataset for training the LLM.
% 4.1 SRA-MCTS
\subsection{SRA-MCTS}
% 我们构建了一种由大模型self完成的基于MCTS的reasoning augment数据生成方法，叫做SRA-MCTS，来生成问题的natural language reasoning plan，并通过质量评估确保plan的正确性。

We develop a self-driven reasoning augmentation method with MCTS, called SRA-MCTS, to generate natural language thinking for questions.
% SRA-MCTS分为四个阶段：Selection，Expansion，Evaluation & Reflection，Backpropagation，这四个阶段被重复多次，每一次重复为一次迭代，生成一个具体推理步骤。大模型在中间两个阶段起到关键作用，而第一阶段和最后阶段由公式或规则来指导进行。
It consists of four phases: \textbf{Selection}, \textbf{Expansion}, \textbf{Evaluation \& Reflection}, and \textbf{Backpropagation}, as shown in Figure~\ref{fig:mcts}.
These phases are carried out on a search tree composed of tree nodes and will be repeated multiple times, with each iteration generating a specific reasoning step.
% The LLM plays a key role in the middle two phases, while the first and last phases are guided by formulas or rules.

% 数据生成流程中的一个tree node有4个关键属性：state，问题和目前生成的plan；action，在State下，大模型决定的plan下一步；reward，对state和action的综合评分，将被用于分支选择；reflection，大模型对于下一步Stepn+1的启发或者结束标志，用于指导大模型下一次action的生成。

A tree node has four attributes:
1) \textit{State} represents the current state, including the question and the steps generated so far; 2) \textit{Action} represents the next step generated based on the current state; 3) \textit{Reward} represents the score generated by LLM based on state and action; 4) \textit{Reflection} guides the generation of the next iteration.
The initial search tree consists of only a root node, with its \textit{state} attribute set to the input question, and the other three attributes are empty.

% In this stage, a tree node is defined as Node = (State, Action, Reward, Reflection). State represents the question and the currently generated plan; Action indicates the next step the LLM decides to take based on the current state; Reward is the combined score of the state and action, used for branch selection; Reflection is the LLM’s guidance or termination signal for the next step, helping to direct the generation of the next action.

\subsubsection{Selection}
The purpose of the selection is to select one step from the model's currently generated steps for the next step generation.
The selection phase involves selecting a leaf node $N_d$ from the search tree, where $d$ represents the depth of the leaf node. This node will be expanded in the expansion phase. 

We use UCB1~\cite{DBLP:journals/ml/AuerCF02}, which is calculated based on the node's \textit{reward} value, to measure the value of each node being selected. The formula is as follows:
\begin{equation}
UCB1 = r + c \cdot \sqrt{\frac {2 \cdot \ln{N}}{n}}
\label{Eq:selection}
\end{equation}
% 其中r为该节点目前方案的奖励分数（在Evaluation中由大模型生成），c是人工设定的探索常数，设定为0.5，N为父节点被访问的次数；n_i为该节点被选择的次数。加号的右边是为了平衡选优和探索两个方面，若去掉右边的探索项，则退化为贪婪选择，多样性就会收到损害。被选择的节点将在下一步用于大模型的方案生成，需要注意的是，在第一次的迭代的选择中只有包含问题的根节点（此时也为叶节点），所以默认选择根节点。
where $r$ is the \textit{reward} value of the current node that is generated by the Evaluation \& Reflection phase in the last iteration, $c$ is the exploration constant set to 0.5, $N$ denotes the number of times the parent node has been selected, and $n$ represents the number of times the current node has been selected. UCB1 is designed to balance exploitation and exploration, represented by the terms on the left and right of the formula respectively. If the exploration term is removed, it degenerates into a greedy selection, which harms diversity.

With the selection phase starting from the root node, the node with the highest UCB1 value at the current depth is selected. Then, the process continues by choosing the child node with the highest UCB1 value among all the child nodes of the selected node. This process continues until a leaf node of the search tree is selected.
Note that during the first iteration, only the root node which is also a leaf node exists, so the root node's UCB1 value does not need to be calculated.

\subsubsection{Expansion}
% 扩展
% 这一阶段的目的是生成新的节点from Selection Phase，即生成下一次的plan，based on the existing steps
% 在上一阶段被选择的叶节点N_d将由大模型进行扩展生成代表新方案的新节点。N的State和Action拼接作为新节点的State：
The purpose of the expansion is to generate the next step based on the existing steps.
The expansion phase involves generating several new child nodes based on the selected node $N_d$, and assigning values to the \textit{state} and \textit{action} attributes of the new nodes. 

% 设计state属性是为了存储模型在当前节点已经生成的steps和question，因此对于来自于同一父节点的子节点来说，他们的state值是一样的。
The \textit{state} attribute is designed to store the input question and steps that the LLM has generated up to the current node. Therefore, for child nodes that originate from the same parent node, their \textit{state} values are identical and referred to $S_{d+1}$, as they share the input question and the same history of step. Specifically, $S_{d+1}$ is formed by concatenating the \textit{state} and \textit{action} attributes of the node $N_d$:
\begin{equation}
S_{d+1} = concatenate(S_d, A_d)
\label{Eq:concatenate}
\end{equation}
where $S_d$ and $A_d$ represent the \textit{state} and \textit{action} of the selected node $N_d$, respectively. 
% The specific concatenate method is shown in Appendix~\ref{sec:concatenate}.

The \textit{action} attribute is designed to store the next step generated based on the current \textit{state} and the \textit{reflection} of the parent node. Since different methods can be used to generate different next steps, the \textit{action} attribute can have multiple distinct values, leading to the creation of different child nodes. Each node represents a possible path that can be made from the current state, allowing for the exploration of various possibilities in the search tree.

% 为了产生具有足够多样性的next step，我们采用sample解码方法进行生成。每生成一个step，都表示the selected node产生了一个新的节点，该新节点连同其父节点构成了一个branch。
To generate diversity in the steps, the expansion phase adopts the sample decoding strategy, where each next step is generated one by one. Each time a new step is generated, it represents a new node created from the selected node $N_d$. 
% 为了防止模型生成重复的next step，我们在输入中添加了模型当前已经生成的next steps。这个过程的公式表示如下
To prevent the LLM from generating duplicate next steps, we add the next steps that have already been generated to the input. This process can be represented by the following formula:
\begin{equation}
A_{d+1}^i = 
\begin{cases}
\begin{aligned}
&LLM(S_{d+1}; R_{d}), & \text{$i = 1$} \\
\\
&LLM(S_{d+1}; A_{d+1}^{1}\\&, ..., A_{d+1}^{i-1}; R_{d}), & \text{$i > 1$}
\end{aligned}
\end{cases} 
    \label{Eq:expansion}
\end{equation}
% A_{d+1}^{i}表示被选择的节点生成的第i个子节点的action值
where $A_{d+1}^{i}$ represents the \textit{action} value of the $i$-th child node at depth $d$+1, $R_d$ represents the \textit{reflection} attribute value of the selected node $N_d$, which is generated in the last iteration.
Adding the already generated steps to the input helps the LLM explore different paths in the search space. 
% The specific prompt is shown in Appendix~\ref{sec:prompts}.

% 考虑到计算开销，我们设定扩展的分支数为3。如果有重复的输出，则进行最大尝试次数为5次的重新生成。在这一个阶段不涉及代码的生成,只生成自然语言形式的推理过程.
To manage computational costs, we set the number of expanded nodes to 3. If a duplicate step is detected, the generation process is repeated, with a maximum of 5 retries At this stage, code generation is not involved, only the natural language reasoning process is generated.

\subsubsection{Evaluation \& Reflection}
% 第三个阶段包括两个部分，这两部分分别作为选择和扩展阶段的支撑。
The third phase consists of two parts: Evaluation and Reflection, which support the selection and expansion phases, respectively.

% Evaluation阶段对节点进行评分，用于对不同分支的节点进行选择扩展。
% 我们使用大模型对所有新生成的节点奖励评分。从以下几个角度进行评判：单步正确性，整体方案连贯性，解决方案完整性，整体方案正确性。若中间步骤出现了错误的步骤，则会获得较低的奖励得分，从而降低该分支被选择的概率，缓解了链式推理因为出现中间错误步骤而导致的鲁棒性损失。奖励评分如下
\paragraph{Evaluation}
% Evaluation部分为Expansion阶段生成的新节点的Reward属性赋值，用来在Selection阶段进行节点选择。
The evaluation part assigns the \textit{reward} attribute values to the new nodes generated in the expansion phase. These reward values will be then used in the selection phase to calculate the UCB1 values.

\begin{figure}[]
    \centering
    \includegraphics[width=\linewidth]{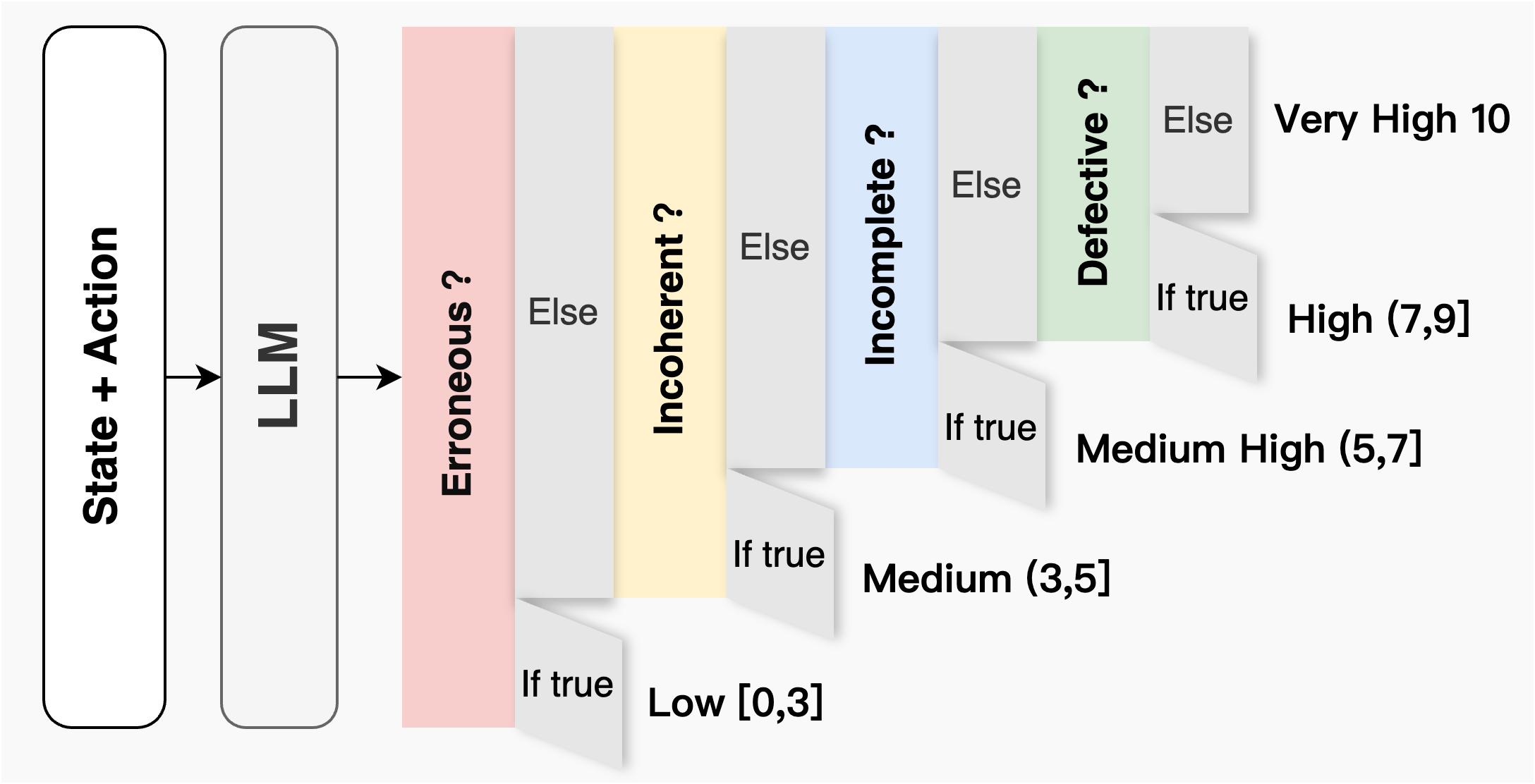}
    \caption{The progressive scoring method. The \textit{state} and \textit{action} of the node are used as inputs, and the judgment is made sequentially from left to right based on the four principles. If the current principle is satisfied, an integer score in the corresponding interval is output; otherwise, the next principle is evaluated. If all the principles are not met, the model will give the current input a full score of 10.}
    % 评分的细则如图，先检查Plan中是否包含错误，再评估Plan是否完整，能够解决问题。 The evaluation criteria, as shown in the diagram, first check whether the $Plan$ contains any errors and then assess whether the $Plan$ is complete and capable of solving the question.
    \label{fig:Evaluation}
\end{figure}

% 我们使用LLM为新节点进行打分
We use the LLM itself, which generates the steps during the expansion phase, as the evaluator to score the new nodes and explore its potential for self-improvement through self-evaluation.~\footnote{Note: The use of the model itself here is aimed at exploring the potential for end-to-end self-improvement. Larger open-source models and closed-source model APIs, which may have more accurate evaluation capabilities, can easily replace this component.} The \textit{reward} value of the $i$-th new node $N_{d+1}^{i}$ is as follows:
\begin{equation}
    r_{d+1}^i = LLM(S_{d+1}^i, A_{d+1}^i)
    \label{Eq:reward}
\end{equation}
% 我们采用渐进式打分方法，从四个方面按顺序逐个进行判断，这四个方面的判断顺序依次是：Single-step correctness, Solution coherence, Solution completeness, and Solution correctness。如过满足当前方面的条件则直接给出对应分数，而则忽略其它方面。
We use a progressive scoring method, making judgments sequentially from four aspects, the process is shown in Figure~\ref{fig:Evaluation}. The order of judgment is as follows: Single-step correctness, Solution coherence, Solution completeness, and Solution correctness. If the criteria for the current aspect are met, the corresponding score is given directly, and the other aspects are ignored.

% and the specific prompt is shown in Appendix~\ref{sec:prompts}.

% 反思
% The Reflection phase aims to guide the reasoning process for generating the next step，i.e.，通过生成一段简短的思路，在Expansion phase拼接在input之后来引导模型的生成
% 在扩展，大模型产生了新节点，但由它们直接作为大模型的输入，生成下一步的步骤效果往往不够稳定。于是我们使用了reflection机制，模型先基于上文，生成一条简短的reflection，提示下一步生成的思路，再由大模型按这个思路生成具体的步骤内容。reflection同样可以用来判断本次生成的步骤是否能够解决这个问题，若判断之前的步骤已经可以正确地解决问题，则输出<end>标签，提示数据生成到此为止。$R = LLM(S_{d+1} + A_{d+1})$取值如下：

\paragraph{Reflection} 
The reflection part aims to guide the next generation direction for the current node, and assigns the \textit{reflection} attribute values to the new nodes generated in the expansion phase.
The \textit{reflection} attribute is a brief thought generated by the LLM and will be used in the next iteration's expansion phase.

We use the reflection~\cite{DBLP:journals/corr/abs-2406-03816} mechanism to allow the model to think before generating the next action, thus preventing performance degradation. The $reflection$ value of the $i$-th new node $N_{d+1}^i$ is as follows:
\begin{equation}
    R_{d+1}^i = LLM(S_{d+1}^i, A_{d+1}^i)
    \label{Eq:reflection-1}
\end{equation} 

Additionally, the \textit{reflection} value is used to assess whether all the generated steps have solved the question. If the generated steps from \textit{state} and \textit{action} at this node are deemed sufficient to solve the question, the \textit{reflection} value will be an $<end>$ tag to indicate that the SRA-MCTS process has concluded. Otherwise, the LLM will generate a reflection for the next iteration's reasoning.
% The specific prompt is shown in Appendix~\ref{sec:prompts}, and 
The \textit{reflection} value is shown as follows:
\begin{equation}
\small
R =
\begin{cases}
    <end>, & \text{question solved} \\[1em]
    \text{short thought for next step}, & \text{question unsolved}
\end{cases}
\label{Eq:reflection-2}
\end{equation}
\normalsize

% 反向传播
% 反向传播是从叶节点向父节点递归传递奖励增量，让下一次迭代的选择环节能更加准确的方式。Plan在初期不一定能得到高得分，但在迭代几次之后或许变得可靠，被选择的概率应该增大，所以通过向上传递子节点增量的方式来实现。形式如下：
\subsubsection{Backpropagation} Backpropagation aims to update the \textit{reward} values of all parent nodes of the current node, making the reward values of the parent nodes more accurate, thereby affecting the generation of the thinking.
% 对于在expansion阶段生成的新子节点，他们共同更新一个父节点，即在selection阶段选中的节点。之后，刚刚的父节点继续更新它的父节点，这个过程一直重复到root节点被更新。
For the new nodes generated during the expansion phase \{${N_{d+1}^{1}, N_{d+1}^{2}, ..., N_{d+1}^I}$\}, they collectively update their parent node, which is the node selected in the selection phase $N_d$. Afterward, the parent node $N_d$ continues to update its parent node $N_{d-1}$, and this process is repeated until the root node is updated. The update process of the parent node's \textit{reward} value $r_p$ is as follows:
\begin{equation}
\begin{aligned}
r_{\Delta} = \sum_{i=1}^{I} (v_{c}^i \  \cdot \ r_{c}^i) / \sum_{i=1}^{I} v_{c}^i  \\
r_{p} = \alpha \cdot r_{p} + (1 - \alpha) \cdot r_{\Delta}  
\end{aligned}
\end{equation}
% 其中Visits_{c}代表节点被访问的次数，用来平衡被多次访问的节点对增量的贡献,$r_{increment}$表示对父节点reward提供的增量，\alpha用于平衡父节点原reward和子节点的increment，我们人工设定为0.7。
where $v_{c}^i$ represents the number of times the $i$-th child node is selected, $r_{c}^i$ represents the \textit{reward} score of the $i$-th child node, $r_{\Delta}$ represents the weighted increment of the parent node's \textit{reward} value, and $\alpha$ is a hyperparameter to balance the original \textit{reward} of the parent node and the increment from the child nodes. 

% 更新后的父节点奖励值在后续的迭代中会影响到子节点的选择。具体来说，在下一个selection阶段，SRA-MCTS会使用更新后的奖励值来评估不同节点的优劣，从而决定探索哪个分支。这种动态奖励调整有助于逐渐调整整个树的策略，使得搜索更加高效，逐步收敛到最优解。
The updated \textit{reward} value of the parent node will influence the selection of child nodes in subsequent iterations. Specifically, in the next selection phase, SRA-MCTS will use the updated \textit{reward} values to evaluate the relative merits of different nodes, thereby deciding which branch to explore. This dynamic reward adjustment helps gradually refine the strategy across the entire tree, making the search more efficient and progressively converging toward the optimal solution.

% 重复上面的四个步骤直到找到正确答案，输出<end>标签，或达到迭代次数的限制。
SRA-MCTS repeats the four phases above until the $<end>$ tag is output, or the iteration limit is reached.

% Code Generation
\subsection{Thinking to Code}
% SRA-MCTS会根据问题产生多个steps，他们构成了一个plan。
SRA-MCTS generates multiple steps based on the question, and these steps together form a thinking.
% 现在我们通过让大模型扮演功能实现员，将Plan严格对应地转换成Code，即$code = LLM(question + plan)$。由SRA-MCTS，我们得到了一批自然语言的Plan，通过下面的prompt使用大模型完成转换：
In this stage, we have the LLM act as a function implementer, strictly converting the thinking into code. 

% Training
\subsection{Training}
% 我们已经收集到Question，Plan，Code三元组，我们将Question作为输入，后面二者作为输出构成训练数据集。由于我们的方法注重自我驱动，如果不特别说明，我们的模型训练的数据集都是自我合成的。在训练上，我们使用LLaMA-Factory进行LoRA fine-tuning。
% After the first two stages, we collect the question, thinking, and code triples, with the question as input and the latter two as output, forming the training dataset. We use the dataset for supervised fine-tuning.
After the first two stages, we collect the question, thinking, and code triples, with the question as the input and the latter two as the output, forming the training dataset. This dataset is then used for supervised fine-tuning.

\section{Experiment}

\subsection{Training Data Generation}
% Note，直觉上大模型的评估能力会比小模型更好，但经过额外方法，小模型也能完成评估任务，尽管上限低于更大模型。这里，我们使用小模型本身使用In-Context Learning进行评估，来探索模型通过自我评估达到自我提升的潜力。
% In the data generation framework's evaluation phase, we use In-context learning for evaluation with the small model itself to explore the potential for self-improvement through self-evaluation. Intuitively, larger models are expected to have better evaluation capabilities than smaller ones. However, with additional methods, smaller models can also carry out evaluation tasks~\cite{DBLP:conf/nips/Ouyang0JAWMZASR22}, although their upper limit is lower than that of larger models. 
% In the evaluation phase of the data generation framework, we use In-context learning for evaluation with the small model itself to explore the potential for self-improvement through self-evaluation. Intuitively, larger models are expected to exhibit better evaluation capabilities than smaller models. However, with additional methods, smaller models can also perform evaluation tasks~\cite{DBLP:conf/nips/Ouyang0JAWMZASR22}, although their upper limit is lower than that of larger models.

% 为了验证SRA-MCTS是否确实能够生成质量更好的thinking，我们需要收集一些具有一定难度的问题。因此我们选择LeetCode dataset中medium和hard难度的数据，而去掉easy难度的数据。我们只保留问题描述而舍弃答案，来让不同方法生成thinking和code。
% 为了防止训练集与测试集出现重复问题，我们对训练集做了去污染处理。方法是对数据集中的每个问题进行10-gram级别的重复检测，阈值设定为0.5，当与测试集的重复度超过阈值时，我们将这条数据从训练集中删除。最终我们得到了1819个问题。
\paragraph{Question} To validate whether SRA-MCTS can indeed generate higher-quality thinking, we need to collect questions with a certain degree of difficulty. Therefore, we select medium and hard-level questions from the LeetCode dataset~\cite{greengerong_leetcode}, excluding easy-level questions. We retain only the question descriptions and discard the answers, allowing different methods to generate both thinking and code.
To prevent any overlap between the training and test sets, we perform decontamination on the training data. Specifically, we conduct 10-gram level duplicate detection on each question in the dataset, setting the threshold at 0.3. If the similarity between a training question and any test set question exceeds this threshold, we remove the corresponding training data. As a result, we obtain a final training set of 1,819 unique questions.

% 对于thinking和code生成，我们在训练集的问题上使用SRA-MCTS以来生成对应的thinking，其中每个step的生成都以0shot方式进行，我们要求模型按照指定的format进行step生成，以便我们提取结果。如果thinking中包含了code，使用正则表达式将其删除从而只保留自然语言推理内容。在生成最终的thinking后，将问题和对应的thinking拼接提示模型来生成最终的code。
\paragraph{Thinking and Code} For thinking and code generation, we use SRA-MCTS to generate the corresponding thinking for the questions in the training set. Each step in the generation process is performed in a zero-shot manner. We prompt the model to generate each step according to a specified format, ensuring that the steps can be easily extracted.
If the thinking contains code, we use a regular expression to remove it, leaving only the natural language reasoning content. This ensures that the reasoning process is kept separate from the code, enabling a clearer focus on the thought process behind the code generation.
After generating the final thinking, we concatenate the question with its corresponding thinking and prompt the model to generate the final code.

\subsection{Baseline} 
% 我们选择gemma-2-2b-it，Meta-Llama-3.1-8B-Instruct和Qwen2.5-14B-Instruct作为我们的backbone模型。选择它们的原因是它们有着良好的指令遵循能力，并且在预训练时已经学习过海量的代码数据，有着通过推理增强方式激活主动思考解决编程问题的潜力。我们直接使用backbone模型对测试集进行推理，来作为性能比较的起始分数。
We choose gemma-2-2b-it~\cite{team2024gemma}, Meta-Llama-3.1-8B-Instruct~\cite{dubey2024llama}, and Qwen2.5-14B-Instruct~\cite{qwen2.5} as our backbone models.
The rationale behind this selection is their strong instruction-following capabilities and the fact that they have been pre-trained on extensive code data. These models demonstrate significant potential to activate proactive thinking for solving programming tasks through reasoning augmentation techniques.
We directly use the backbone models to perform reasoning on the test set, which serves as the baseline performance score for comparison.

% 我们将CoT和ToT推理方法应用于每个模型作为baseline方法。对于CoT，采用传统的提示“让我们一步一步思考”来生成thinking。对于ToT，我们使用深度优先搜索来生成thinking，最大深度设置为4，分支因子设置为3，这与SRA-MCTS中expansion阶段的扩展的节点数量一致。同SRA-MCTS一样，这些thinking也会经过正则处理来排除掉代码内容。最后，将这些方法生成的thinking与question拼接起来输入到模型来得到最终的code。
We apply the Chain-of-Thought (CoT)~\cite{DBLP:conf/nips/Wei0SBIXCLZ22} and Tree-of-Thought (ToT)~\cite{DBLP:conf/nips/YaoYZS00N23} reasoning methods to each model as baseline methods. For CoT, we use the conventional prompt, ``Let's think step by step'', to generate the thinking. For ToT, we use depth-first search to generate the thinking, setting the maximum depth to 4 and the branching factor to 3, which aligns with the number of nodes expanded during the expansion phase of SRA-MCTS.
Similar to SRA-MCTS, the thinking generated by these methods is processed using regular expressions to remove any code content. Finally, the generated thinking is concatenated with the question and fed into the model to produce the final code.

\subsection{Test Set}
% 为了评测方法的效果，我们使用在代码领域中常用的benchmark：Human-Eval和MBPP。此外，为了更准确的评估，我们还使用了EvalPlus框架，其中的HumanEval+和MBPP+的测试样例是原版的数十倍。其中，Human-Eval系的benchmark为续写某个函数；MBPP系要求模型从头给出函数。另外，为了验证推理增强的模型的能力，我们使用GPT-4o对Human-Eval和MBPP进行了难度划分，将它们分为Easy，Medium，Hard三个等级，将其中的Medium和Hard部分组成Human-Eval-C和MBPP-C benchmark。
% We use commonly adopted benchmarks in the code generation field, Human-Eval~\cite{DBLP:journals/corr/abs-2107-03374} and MBPP~\cite{DBLP:journals/corr/abs-2108-07732}. Additionally, we also use the HumanEval+ and MBPP+ in EvalPlus~\cite{DBLP:conf/nips/LiuXW023} framework, where the test cases are several times larger than the original version. Among these, the Human-Eval benchmark involves completing a given function, while the MBPP benchmark requires the model to generate the function from scratch. Additionally, to evaluate the capabilities of reasoning-augmented models, we used GPT-4o to filter the difficulty levels of Human-Eval and MBPP, selecting the relatively challenging problems to create the Complex splits of Human-Eval and MBPP.~\footnote{Specifically, we use GPT-4o to classify the questions into three categories of easy, medium, and hard. The easy category is discarded, while the remaining two categories are retained to create the Complex split.} The same process was applied to Human-Eval+ and MBPP+.
\paragraph{Benchmark}
\label{benchmark}
We use commonly adopted benchmarks in the code generation field, including Human-Eval~\cite{DBLP:journals/corr/abs-2107-03374} and MBPP~\cite{DBLP:journals/corr/abs-2108-07732}. Additionally, we utilize Human-Eval+ and MBPP+ within the EvalPlus~\cite{DBLP:conf/nips/LiuXW023} framework, where the test cases are several times larger than the original versions. As for the task type, Human-Eval involves completing a given function, while MBPP requires the model to generate the function from scratch. 

% 为了评估模型在不同难度数据集上的推理能力，我们对测试集的问题进行了难度划分。具体来说，我们使用GPT-4o将问题分为easy，medium和hard三种难度，全部难度数据组成了Full split，medium和hard数据组成了complex split。

\paragraph{Test Set Split} To evaluate the model's reasoning ability across different difficulty levels, we categorize the test set questions into different difficulty levels. Specifically, we use GPT-4o~\cite{hurst2024gpt} to classify the questions into easy, medium, and hard categories. All questions across the three difficulty levels form the Full split and the medium and hard questions are grouped into the Complex split.

% 在代码生成领域，Pass@k是常用的评估指标，用于衡量模型在k次尝试中至少生成一次正确代码的概率。Pass@k能反映模型在多次生成中的成功率，特别适用于代码生成领域，因为生成过程具有不确定性。pass@k的详细定义和公式可以参考附录

% In code generation, pass@k is a common metric that measures the probability of generating a correct solution in at least one of $k$ attempts~\cite{DBLP:journals/corr/abs-2107-03374,DBLP:conf/nips/KulalPC0PAL19}. It reflects the model's success rate across multiple attempts, which is especially useful given the uncertainty in the generation process. 
\paragraph{Evaluation Metric} In code generation field, pass@k is a widely used metric that measures the probability of generating a correct solution in at least one of $k$ attempts~\cite{DBLP:journals/corr/abs-2107-03374,DBLP:conf/nips/KulalPC0PAL19}. It reflects the model's success rate across multiple attempts, which is particularly valuable given the inherent uncertainty in the generation process.
% The detailed definition and formula for pass@k can be found in the Appendix~\ref{sec:pass@k}.

% \usepackage{graphicx}
\begin{table*}[]
\centering
\resizebox{\textwidth}{!}{%
\begin{tabular}{lrrrrrrrrrrr}
\toprule
 &
  \multicolumn{2}{c}{\textbf{MBPP}} &
  \multicolumn{3}{c}{\textbf{MBPP+}} &
  \multicolumn{2}{c}{\textbf{Human-Eval}} &
  \multicolumn{3}{c}{\textbf{Human-Eval+}} &
  \multicolumn{1}{c}{\textbf{Average}} \\  \cmidrule(lr){2-3} \cmidrule(lr){4-6} \cmidrule(lr){7-8} \cmidrule(lr){9-11}
 \textbf{Method} &
  \multicolumn{2}{c}{pass@1} &
  \multicolumn{2}{c}{pass@1} &
  \multicolumn{1}{c}{pass@10} &
  \multicolumn{2}{c}{pass@1} &
  \multicolumn{2}{c}{pass@1} &
  \multicolumn{1}{c}{pass@10} &
  \textbf{} \\ \cmidrule(lr){2-3} \cmidrule(lr){4-5} \cmidrule(lr){6-6} \cmidrule(lr){7-8} \cmidrule(lr){9-10} \cmidrule(lr){11-11}
 &
  \multicolumn{1}{c}{Full} &
  \multicolumn{1}{c}{Complex} &
  \multicolumn{1}{c}{Full} &
  \multicolumn{1}{c}{Complex} &
  \multicolumn{1}{c}{Full} &
  \multicolumn{1}{c}{Full} &
  \multicolumn{1}{c}{Complex} &
  \multicolumn{1}{c}{Full} &
  \multicolumn{1}{c}{Complex} &
  \multicolumn{1}{c}{Full} &
  \textbf{Increment} \\ \midrule
\multicolumn{12}{c}{\textbf{gemma-2-2b}} \\ \midrule
Instruct &
  34.42 &
  11.32 &
  43.39 &
  \multicolumn{1}{r}{13.64} &
  48.41 &
  39.76 &
  25.42 &
  33.23 &
  \multicolumn{1}{r}{18.64} &
  51.22 &
  - \\ \hdashline
CoT &
  34.90 (+0.48) &
  11.32 (+0.00) &
  43.70 (+0.31) &
  13.64 (+0.00) &
  47.90 (-0.51) &
  41.89 (+2.13) &
  25.42 (+0.00) &
  34.94 (+1.71) &
  20.34 (+1.70) &
  53.05 (+1.83) &
  +0.77 \\
ToT &
  33.86 (-0.56) &
  9.43 (-1.89) &
  44.71 (+1.32) &
  9.09 (-4.55) &
  47.62 (-0.79) &
  40.18 (+0.42) &
  23.73 (-1.69) &
  32.68 (-0.55) &
  20.34 (+1.70) &
  45.12 (-6.10) &
  -1.27 \\
SRA-MCTS &
  33.92 (-0.50) &
  \textbf{16.98 (+5.66)} &
  45.37 (+1.98) &
  \textbf{15.90 (+2.26)} &
  49.21 (+0.80) &
  40.73 (+0.97) &
  \textbf{25.42 (+0.00)} &
  34.88 (+1.65) &
  \textbf{20.34 (+1.70)} &
  49.39 (-1.83) &
  \textbf{+1.27} \\ \midrule
\multicolumn{12}{c}{\textbf{Meta-Llama-3.1-8B}} \\ \midrule
Instruct &
  51.94 &
  33.96 &
  45.37 &
  \multicolumn{1}{r}{29.55} &
  74.60 &
  62.74 &
  \textbf{47.46} &
  58.90 &
  \multicolumn{1}{r}{\textbf{40.68}} &
  67.68 &
  - \\ \hdashline
CoT &
  52.94 (+1.00) &
  39.62 (+5.66) &
  60.50 (+15.13) &
  \textbf{40.91 (+11.36)} &
  74.60 (+0.00) &
  62.32 (-0.42) &
  28.81 (-18.65) &
  58.35 (-0.55) &
  22.03 (-18.65) &
  66.46 (-1.22) &
  -0.63 \\
ToT &
  52.72 (+0.78) &
  32.08 (-1.88) &
  60.24 (+14.87) &
  31.82 (+2.27) &
  74.07 (-0.53) &
  62.26 (-0.48) &
  40.68 (-6.78) &
  57.44 (-1.46) &
  32.20 (-8.48) &
  63.41 (-4.27) &
  -0.60 \\
SRA-MCTS &
  54.52 (+2.58) &
  \textbf{45.28 (+11.32)} &
  59.97 (+14.60) &
  38.64 (+9.09) &
  75.66 (+1.06) &
  62.19 (-0.55) &
  44.07 (-3.39) &
  57.87 (-1.03) &
  38.98 (-1.70) &
  68.29 (+0.61) &
  \textbf{+3.26} \\ \midrule
\multicolumn{12}{c}{\textbf{Qwen2.5-14B}} \\ \midrule
Instruct &
  56.42 &
  56.60 &
  61.48 &
  \multicolumn{1}{r}{52.27} &
  70.37 &
  80.37 &
  69.49 &
  76.52 &
  \multicolumn{1}{r}{61.02} &
  90.95 &
  - \\ \hdashline
CoT &
  58.12 (+1.70) &
  45.28 (-11.32) &
  63.97 (+2.49) &
  31.82 (-20.45) &
  70.37 (+0.00) &
  78.66 (-1.71) &
  61.02 (-8.47) &
  73.84 (-2.68) &
  54.24 (-6.78) &
  90.24 (-0.71) &
  -4.79 \\
ToT &
  59.98 (+3.56) &
  54.72 (-1.88) &
  62.67 (+1.19) &
  45.45 (-6.82) &
  70.63 (+0.26) &
  75.61 (-4.76) &
  64.41 (-5.08) &
  73.84 (-2.68) &
  54.24 (-6.78) &
  90.24 (-0.71) &
  -2.37 \\
SRA-MCTS &
  64.20 (+7.78) &
  \textbf{71.70 (+15.10)} &
  61.16 (-0.32) &
  \textbf{63.64 (+11.37)} &
  83.60 (+13.23) &
  85.37 (+5.00) &
  \textbf{69.49 (+0.00)} &
  75.00 (-1.52) &
  \textbf{61.02 (+0.00)} &
  91.46 (+0.51) &
  \textbf{+5.12} \\ \bottomrule
\end{tabular}%
}
\caption{Main results of different methods. ``Instruct'' represents the method of directly using the backbone model for inference. ``CoT'', ``ToT'', and ``SRA-MCTS'' represent the methods of using the data generated by the corresponding methods for training. ``Full'' represents the entire dataset, and ``Complex'' represents the dataset consisting of the medium and hard categories classified by GPT-4o.}
\label{tab:main}
\end{table*}

\subsection{Training Setup}
% 在SRA-MCTS中，总迭代次数限制为5，Expansion phase中模型的生成temperature我们设定为0.9，并且设定top_p=0.98的采样，在Evaluation phase，我们选择模型自己作为评估者。
% In SRA-MCTS, the total iteration limit is set to 5, the generation temperature in the expansion phase is set to 0.9, with a top-p sampling value of 0.98, and the $\alpha$ in the backpropagation phase is set to 0.5 manually. 
In SRA-MCTS, the total iteration limit is set to 5, the generation temperature in the expansion phase is set to 0.9, with a top-p sampling value of 0.98, and the $\alpha$ value in the backpropagation phase is manually set to 0.5.

% 我们使用baseline中提到的模型作为训练的基座，learning_rate设置为1.0e-4，训练2个epoch，warmup_ratio为0.05，batch size为4。
We use the models mentioned in the baseline as the foundation for training, setting the learning rate to 1e-4, training for 2 epochs, with a warmup ratio of 0.05 and a batch size of 4. 
% 对于每个量级的模型，我们使用4种来源的数据集分别进行训练：1）Meta-Llama-3-70B-Instruct使用CoT蒸馏的数据；2）模型自己使用CoT生成的数据；3）模型自己使用ToT以DFS生成的数据；4）模型自己使用SRA-MCTS生成的数据.
For each model scale, we train using three different datasets: 1) Self-generated CoT data; 2) Self-generated ToT data; 3) SRA-MCTS generated data. 
% 我们使用Llama-factory进行LoRA微调，2B，8B，14B模型分别使用2张TITAN RTX，RTX 3090，RTX A6000进行训练。
We use Llama-factory~\cite{zheng2024llamafactory} for LoRA~\cite{DBLP:conf/iclr/HuSWALWWC22} fine-tuning, with the models trained on two RTX A6000 GPUs.

\subsection{Main Results}

% 【translate-done】 SRA-MCTS整体性能最佳，并且在解决复杂问题方面表现出色。正如表所示，我们的方法在所有模型上展示出了最高的平均性能提升，而且在大多数benchmark的complex split上的提升幅度要显著大于full split，尤其在MBPP上有大约2-15个点的增加。然而对于Human-Eval相关的benchmark，所有的方法都出现很多指标下降的情况，我们的方法在部分指标上仍有提升但与其它baseline方法相比并不明显。我们将这个现象归因与训练集数据与测试集数据的任务类型在巨大差异。正如在section\ref{benchmark}中提到的任务类型说明，MBPP和训练集一样要求模型根据给定的问题从头生成对应的代码，而Human-Eval要求模型按照注释续写给定的函数头。因此基于MBPP与训练集风格的一致性，我们的方法相比其它方法能够产生更大的提升。这也说明了训练集的数据风格会对下游任务产生重要的影响。
\paragraph{SRA-MCTS demonstrates the best overall performance and excels in solving complex questions.} As shown in Table~\ref{tab:main}, our method achieves the highest average performance improvement across all models, with a significantly larger gain on the Complex split of most benchmarks compared to the Full split, particularly with an increase of approximately 2-15 points on MBPP and MBPP+. However, for the Human-Eval-related benchmarks, all methods experience a decline in several metrics. While our method still shows improvement in some metrics, it is not obvious compared to the other baseline methods. 
We attribute this phenomenon to the differences between the task types in the training and test sets. As discussed in Section \ref{benchmark}, tasks in the MBPP benchmark, like those in the training set, require the model to generate code from scratch based on a given question, whereas Human-Eval tasks ask the model to complete a function header provided in the prompt. Due to the consistency in task style between MBPP and the training set, our method achieves more substantial improvements compared to other methods. This highlights the critical role that the style of training data plays in influencing downstream task performance.

\paragraph{The larger the model, the more pronounced the improvement from our method.} As we observe in Table~\ref{tab:main}, the average performance improvement increases with the model size, and this trend is even more evident on the Complex split. We attribute this to the fact that larger models possess better generation and evaluation capabilities. With stronger generation and evaluation capabilities enabled by larger model scales, these models can produce higher-quality data and more effectively identify errors and redundancies in reasoning steps. SRA-MCTS effectively leverages these enhanced capabilities, leading to significant performance gains. 

% 我们方法的提升效果会随着模型的自我评估能力的提高而增强
% 在主表中,我们可以明显地发现随着模型量级的增大,平均提升量也在增大,我们将这归因为更大模型有着更强的评估能力,模型在Complex split上的提升量更加佐证了这一趋势,在MBPP+ Complex split上三个量级有着翻倍的提升.随着模型评估能力的增强,它们能更好地识别推理步骤中的错误和冗余,从来带来更高质量的数据,这一点在后面的数据分析实验中也得到了验证.所以我们认为,当实验推广到更大的模型,如33B,70B时,我们的方法会有更明显的提升.
% \paragraph{The effectiveness of our method improves as the model's self-generation and self-evaluation capabilities strengthen.} It is clearly observed in Table~\ref{tab:main} that as the model scale increases, the average improvement also grows. We attribute this to the larger models' enhanced generation and evaluation capabilities. This trend is further supported by the improvements observed on the Complex splits, particularly on the MBPP+ Complex split, where the improvement doubles across the three model scales. With stronger generation and evaluation capabilities enabled by larger model scales, these models can produce higher-quality data and more effectively identify errors and redundancies in reasoning steps. This observation is further validated by subsequent data analysis experiments. Therefore, we anticipate that our method will yield even greater improvements when applied to larger models.

\section{Analysis}
\subsection{Ablation Study of the Thinking}

\begin{figure}
    \centering
    \includegraphics[width=\linewidth]{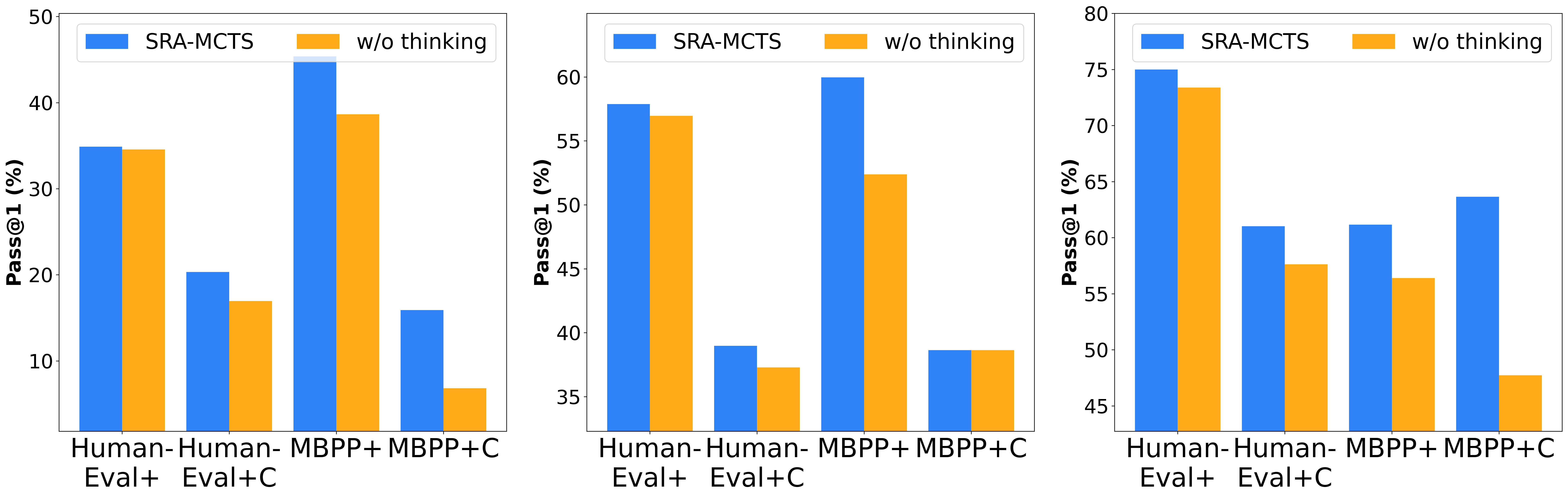}
    \caption{Comparison results of including and excluding thinking in the training data. ``w/o thinking'' represents the model is trained without thinking in the training set. ``C'' represents the Complex split.} 
    % 这张图比较了有无thinking对模型各方面性能的影响.w/o thinking代表去除了训练集中thinking后重训练的模型.
    \label{fig:ablation}
\end{figure}

% 【translate-done】thinking存在的影响。我们额外训练了一个不带有thinking，仅包含问题和code的数据集来与包含三元组的数据集进行比较。为了公平比较，该数据集的code直接从由SRA-MCTS产生的训练集中提取。结果如图所示，几乎所有不带有thinking的结果都要低于带有thinking的，这个差距在complex split上更加明显。我们认为这可能是由于太简单的问题不需要太多的思考模型就能给出正确的解决方案，正是由于问题难度增加，模型才需要更高质量的thinking来引导出正确的code。
\paragraph{Effect of Existence} We additionally train a model on a dataset without thinking, containing only the question and code, and compare it with the dataset that includes the question, thinking, and code triples. For a fair comparison, the code in the dataset without thinking is directly extracted from the training set generated by SRA-MCTS. 
As shown in Figure~\ref{fig:ablation}, nearly all cases without thinking perform worse than those with thinking, with this gap being more obvious on the Complex split. We believe this is due to simpler questions not requiring much reasoning from the model to provide the correct solution, allowing a code-only training set to solve these questions effectively. It is only when the question difficulty increases that the model benefits from higher-quality thinking to guide it toward generating the correct code. 

\begin{figure}
    \centering
    \includegraphics[width=\linewidth]{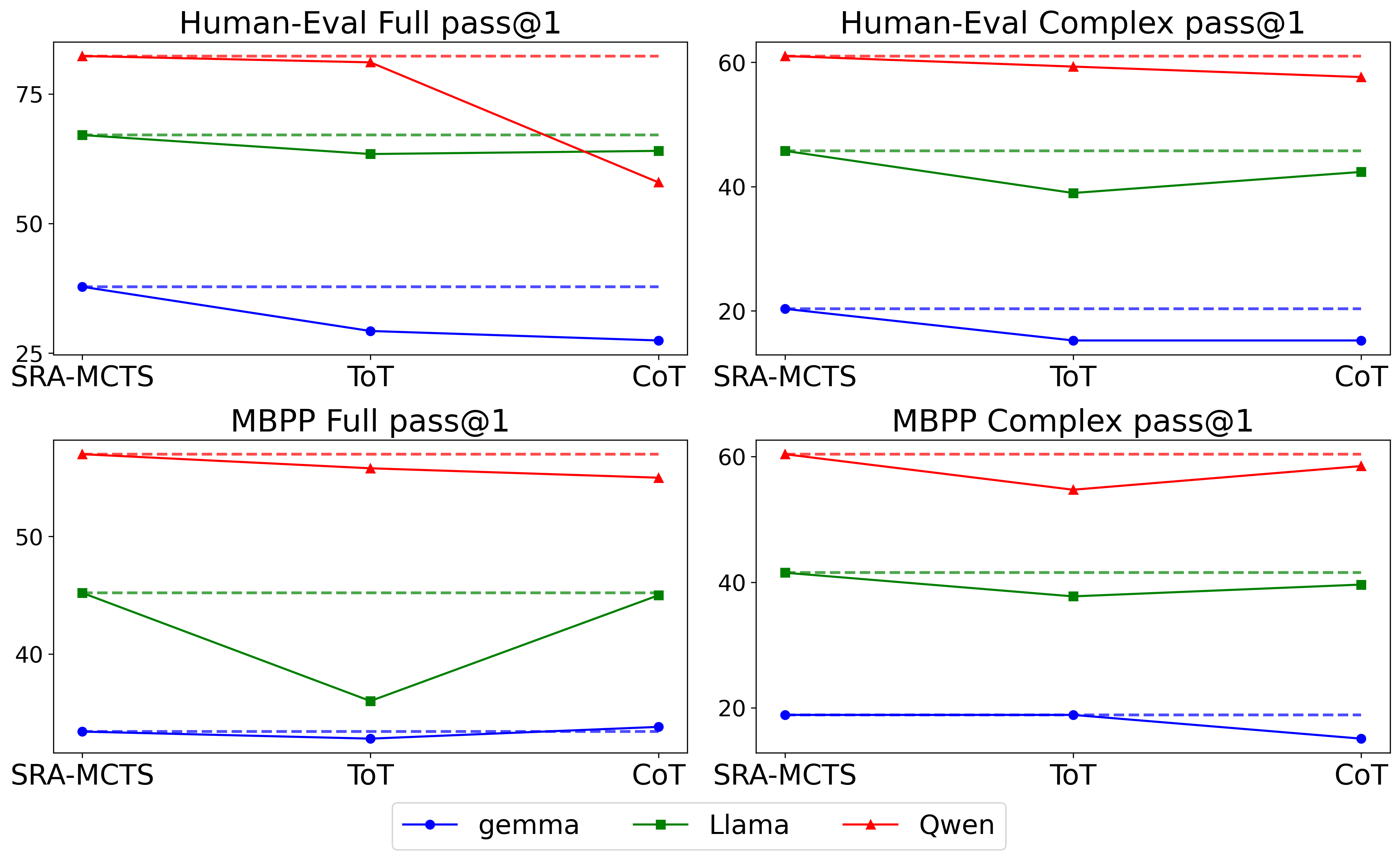}
    \caption{Comparison results of different thinking variants. The dashed line represents the performance of SRA-MCTS.}
% Ablation Study: 研究回答中的自然语言推理步骤的作用，其中SRA-MCTS为按pipeline流程训练的，训练数据包括自然语言推理和代码，另一组为训练数据中的回答只包括代码. CoT's plan和ToT's plan代表用CoT或ToT生成的plan替换SRA-MCTS生成的plan，但保留SRA-MCTS数据中的code部分
    \label{fig:counterpart}
\end{figure}
% \paragraph{The presence of thinking has a limited impact on performance in standard benchmarks but makes a significant difference in complex benchmarks.} To verify the impact of the existence of thinking on model performance, we fine-tune each of the three models referenced in the baseline using two different datasets, resulting in two fine-tuned versions of each model. One dataset consists of the training set generated by SRA-MCTS, which includes both the thinking and code portions, while the other dataset is formed by removing the thinking portion and retaining only the code from the previous dataset. We compare their performance in Figure~\ref{fig:ablation}, models without thinking experience a significant performance gap on the Complex benchmarks, while the gap on the standard benchmarks is less pronounced by comparison. We attribute this to the fact that simple problems do not require detailed decomposition for correct answers, allowing a code-only training set to solve these problems effectively.

% 【translate-done】thinking差异的影响。我们从CoT，ToT和SRA-MCTS产生的训练集中分别提取出对应的thinking，并将其与问题和ground truth代码拼接在一起，形成了三个只有thinking不同的训练集。我们在这个新的训练集上重新训练backbone模型，结果如图所示。我们的方法生成的thinking在三种方法中有着最好的稳定性，这表现在新数据集的code虽然是正确的，但与thiking之前可能存在不一致性。我们的方法在替换code后仍能保持较好的性能，而其他方法都出现了较为明显的性能下降。这说明我们方法生成的thinking不仅能够带来最终的性能提升，而且具有一定的鲁棒性使得模型表现更稳定。
\paragraph{Effect of Variants} We extract the thinkings generated by CoT, ToT, and SRA-MCTS from their respective training sets and concatenate them with the question and ground truth code to create three new training sets, each differing only in the thinking content. We then retrain the backbone models on these new datasets, and the results are shown in Figure~\ref{fig:counterpart}. 
Our method generates the most stable thinking among the three approaches. This is demonstrated by the fact that even when the code is replaced, our method maintains strong performance, whereas the other methods experience noticeable performance degradation. This indicates that the thinking generated by our method not only contributes to the final performance improvement but also provides robustness, leading to more stable model.

% 对于每个模型,我们提取出三种方法生成的thinking,与数据集中的代码官方解拼接在一起,形成三个不同的数据集用于fine-tune模型,在Figure 5中比较它们的性能. 我们的方法生成的thinking在三种方法中有着最好的稳定性,表现在更换了由thinking生成的code后,依然能表现出良好的性能,与其他方法相比有着更稳定且优异的性能.这种稳定性在Table 2中也有体现.说明我们的方法能够在一个更高的视角分析问题,而不会局限于具体的某种方法,所以在拼接其他代码时也能有优秀的表现.
% \paragraph{The thinking generated by SRA-MCTS demonstrates the highest robustness among all methods.}  For each model, we extract the thinking produced by the three aforementioned methods and combine it with the official Python code solutions from the LeetCode dataset to create three distinct datasets for fine-tuning. Their performance is compared in Figure~\ref{fig:counterpart}, where the thinking generated by our method shows the greatest robustness, maintaining strong performance even when its associated code is replaced. Compared to other methods, it delivers more superior performance, a trend also observed in Table~\ref{tab:main}. This robustness suggests that our method is capable of analyzing problems from a higher-level perspective, rather than being confined to a specific approach that may experience performance degradation when integrated with other's code.

\begin{figure}
    \centering
    \includegraphics[width=\linewidth]{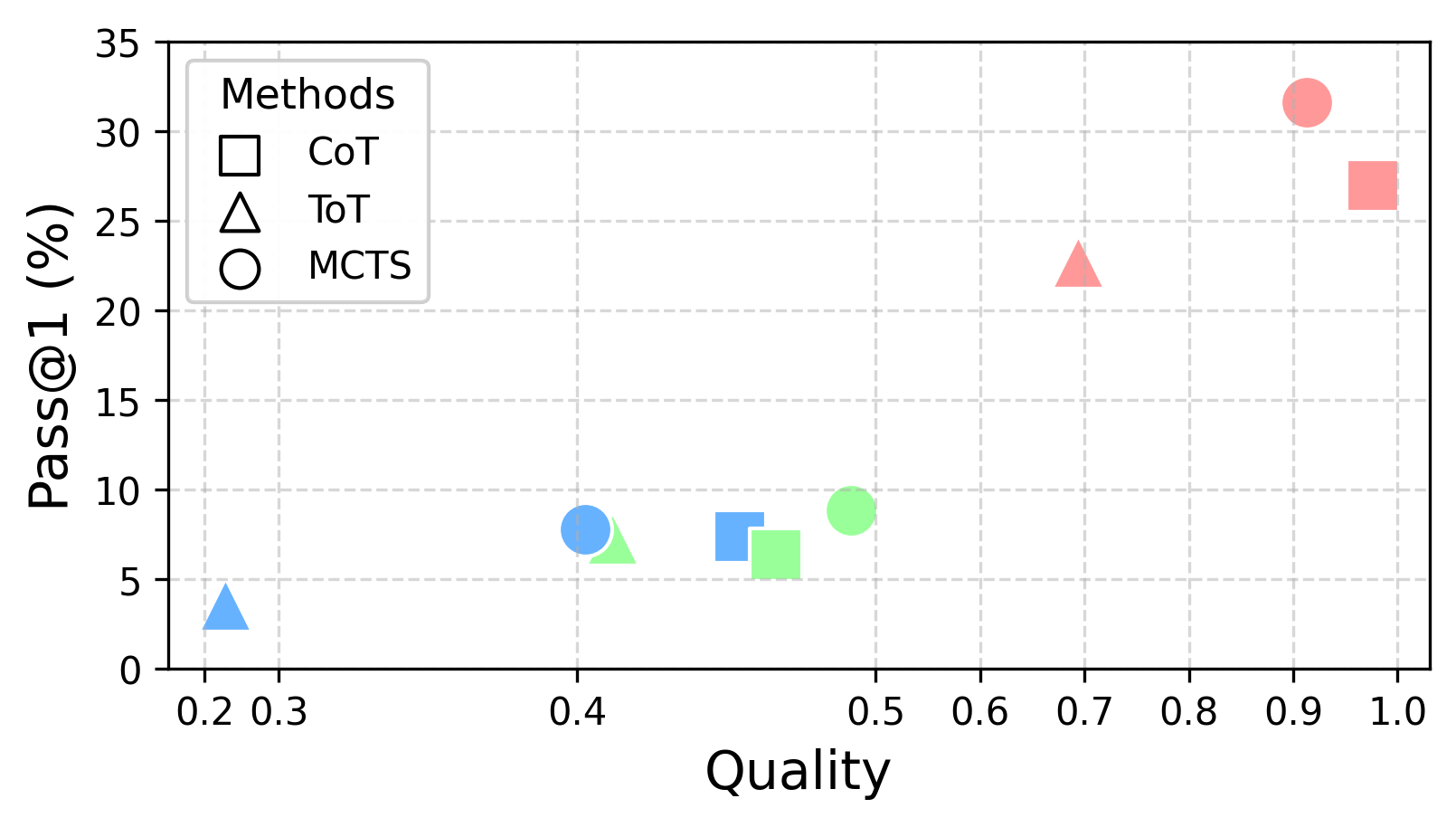}
    \caption{The relationship between the quality of thinking and the pass@1 rate of training set code. \textcolor{red}{Red} represents the Qwen model, \textcolor{green}{green} represents the Llama model, and \textcolor{blue}{blue} represents the gemma model.}
    \label{fig:scatter}
\end{figure}
% 【translate-done】thinking质量的影响。我们使用GPT-4o对thinking的质量进行评估，将每个thinking划分为Low，Medium和High三类，每个训练集中Medium和High所占的比例用来代表这个数据的质量分数。结果如图所示，可以看到训练集thinking的质量与性能大致呈现出线性关系，也就是说thinking的质量越高，模型最终性能也会越好。但是对于我们的方法，在gemma和Qwen模型上也存在质量偏低但性能却偏高的情况。这也一方面说明了大模型的评估不一定可靠，另一方面也说明了我们方法的优越性。
\paragraph{Effect of Quality} We use GPT-4o to evaluate the quality of the thinking, categorizing each instance into low, medium, and high quality. The ratio of medium and high quality thinking in each training set are used to represent the quality score of the dataset. 
As shown in Figure~\ref{fig:scatter}, there is generally a linear relationship between the quality of the thinking in the training set and the model's performance, meaning that higher-quality thinking leads to better final performance. However, for our method, there are cases in the gemma and Qwen models where the thinking quality is relatively low, yet the performance is higher than expected. We hypothesize that, during the evaluation of thinking, the evaluator may overemphasize unnecessary overthinking steps, complicating the model's ability to generate code based on overly complex reasoning processes. This not only suggests that the evaluation of larger models may not always be reliable, but also highlights the superiority of our method. 
\begin{table}[]
\centering
\resizebox{0.98\columnwidth}{!}{%
\begin{tabular}{lcccc}
\toprule
\multirow{2}{*}{\textbf{Model}} &
  \multirow{2}{*}{\textbf{Size}} &
  \multirow{2}{*}{\textbf{MBPP+}} &
  \multirow{2}{*}{\textbf{Human-Eval+}} &
  \multirow{2}{*}{\textbf{Average}} \\
                    &                          &                &                &                \\ \midrule
CodeGen             & 2B                       & 36.00          & 22.60          & 29.30          \\
Codegemma           & 2B                       & \textbf{46.60} & 20.70          & 33.65          \\
SRA-MCTS (Ours)      & 2B                       & 45.37          & \textbf{34.88} & \textbf{40.13} \\ \midrule
DolphCoder          & 7B                       & 52.60          & 54.90          & 53.75          \\
CodeLlama           & 7B                       & 45.40          & 34.10          & 39.75          \\
WizardCoder         & 7B                       & 49.50          & 45.10          & 47.30          \\
Magicoder-S-CL      & 7B                       & \textbf{60.10} & \textbf{67.70} & \textbf{63.90} \\
SRA-MCTS (Ours)      & 8B                       & 59.97          & 57.87          & 58.92          \\ \midrule
DolphCoder          & 13B                      & 54.10          & 57.90          & 56.00          \\
CodeLlama           & 13B                      & 50.90          & 36.60          & 43.75          \\
WizardCoder         & 13B                       & 54.20          & 50.60          & 52.40          \\
StarCoder2-Instruct & 15B                      & \textbf{61.20} & 63.40          & 62.30          \\
SRA-MCTS (Ours)      & 14B                      & 61.16          & \textbf{75.00} & \textbf{68.08} \\ \bottomrule
\end{tabular}%
}
\caption{Comparison the pass@1 results of our method and other competitive methods. The results of other methods are derived from the corresponding papers or the evaluation results of the EvalPlus framework.}
\label{tab:comparison}
\end{table}

\subsection{Further analysis}
% 【translate-done】与有竞争力的同类方法比较。我们同其它具有相同量级模型的方法进行了比较，结果如表所示。对于2B和13B量级的模型来说，我们的方法不仅在单项指标有突出表现，并且在平均上也表现的最好。只有与8B的Magicoder-S-CL方法相比，我们的方法是落后的。我们认为这是由于我们的微调数据数量和多样性上远远低于Magicoder导致的。Magicoder的数据集规模大约是我们的40倍，在大量数据的前提前多样性也很可能要远多于我们的数据集。这也能够体现出我们的方法仅仅使用很小的一部分数据集就能产生足够高质量和高多样性的结果。
\paragraph{Comparison with Competitive Methods.} We compared our approach with other competitive methods using models of similar sizes, as shown in Table~\ref{tab:comparison}. For 2B and 13B models, our method not only excels in individual metrics but also achieves the best overall average performance. The only exception is the 8B Magicoder-S-CL method, where our approach falls slightly behind. We attribute this to the fact that our fine-tuning dataset is significantly smaller and less diverse compared to Magicoder. The Magicoder dataset is approximately 41 times larger than ours, and its larger scale and potential diversity likely contribute to its superior performance. 
This comparison underscores the efficiency of our approach, despite using a small dataset, we are able to generate results that are of sufficiently high quality and diversity. This highlights the effectiveness of our method, which can achieve competitive results even with much more limited data.

% 在与开源同量级模型的比较中，SRA-MCTS也表现出了竞争力。在2B组和13B组中不仅在单项指标有突出表现，并且在平均上也表现的最好。但是在8B中落后于Magicoder，我们这是认为我们的微调数据数量上远远低于导致的，Magicoder的数据集规模大约是我们的37倍。实验表明，我们使用的常规语言模型进行微调，和经过代码数据再训练或从头使用代码数据预训练的模型有着比拟甚至更好的表现。
% \paragraph{In comparison with open-source models of similar scale, SRA-MCTS also demonstrates strong competitiveness.} As shown in Table~\ref{tab:comparison}, our method excels in specific metrics and achieves the best overall performance on average in both the 2B and 13B groups. However, in the 8B group, it falls behind Magicoder. We attribute this gap to the significantly broader scope and substantially larger fine-tuning dataset used by Magicoder, which is approximately 37 times larger than ours and includes diverse instruction formats, such as question-answering, error correction, and completion related to Human-Eval. This discrepancy largely accounts for the performance gap on Human-Eval+. Experimental results indicate that our approach, fine-tuned with limited data, matches or even surpasses models fine-tuned on massive code datasets or pre-trained from scratch.

% 【translate-done】Self-generation的有效性。我们使用Meta-Llama-3-70B作为外部模型来生成训练集，生成的数据用于训练小模型，比较训练数据由自我生成和数据蒸馏这两种方法的性能差异。结果如表所示，我们self-generation的方法在全部大小量级的模型和几乎所有的benchark上超越了数据蒸馏的方法。并且SRA-MCTS在Human-Eval和Human-Eval+这两个数据集上有着可观的提升。而蒸馏方法仅在使用14B模型的MBPP上能够超越。这个结果充分说明了使用小模型自生成数据进行训练的可行性。
\paragraph{Effectiveness of Self-generation.} We use Meta-Llama-3-70B~\cite{dubey2024llama} as an external model to generate a training dataset, which is then used to train smaller models. We compare the performance of models trained with self-generated data versus those trained with data distilled from the larger model. The results in Table~\ref{tab:self-gen} demonstrate that our self-generation approach outperforms the data distillation method across all model sizes and nearly all benchmarks. 
Notably, SRA-MCTS shows substantial improvements on the Human-Eval and Human-Eval+ datasets. In contrast, the distillation method only outperforms on the MBPP benchmark when using a 14B model. This result strongly supports the feasibility of using small models to self-generate data for training, demonstrating that self-generated data can achieve competitive performance without relying on large-scale data distillation.

\begin{table}[]
\centering
\resizebox{\columnwidth}{!}{%
\begin{tabular}{lcccc}
\toprule
\multicolumn{1}{l}{\multirow{2}{*}{\textbf{Model}}} &
  \multirow{2}{*}{\textbf{MBPP}} &
  \multirow{2}{*}{\textbf{MBPP+}} &
  \multirow{2}{*}{\textbf{Human-Eval}} &
  \multirow{2}{*}{\textbf{Human-Eval+}} \\
\multicolumn{1}{l}{}             &                        &                        &                        &                        \\ \midrule
\multicolumn{5}{c}{\textbf{gemma-2-2b}}                                                                                               \\ \midrule
\multicolumn{1}{l}{Distillation} & 34.76                  & 41.93                  & 38.41                  & 33.78                  \\
\multicolumn{1}{l}{SRA-MCTS}     & \textbf{34.88} (+0.12) & \textbf{45.37} (+3.44) & \textbf{40.73} (+2.32) & \textbf{33.92} (+0.14) \\ \midrule
\multicolumn{5}{c}{\textbf{Meta-Llama-3.1-8B}}                                                                                        \\ \midrule
\multicolumn{1}{l}{Distillation} & 56.03                  & 58.09                  & 59.45                  & 51.10                  \\
\multicolumn{1}{l}{SRA-MCTS}     & \textbf{57.87} (+1.84) & \textbf{59.97} (+1.88) & \textbf{62.19} (+2.74) & \textbf{54.52} (+3.42) \\ \midrule
\multicolumn{5}{c}{\textbf{Qwen2.5-14B}}                                                                                              \\ \midrule
\multicolumn{1}{l}{Distillation} & \textbf{71.20}         & 58.65                  & 83.41                  & 73.35                  \\
\multicolumn{1}{l}{SRA-MCTS}     & 64.20 (-7.00)          & \textbf{61.16} (+2.51) & \textbf{85.37} (+1.96) & \textbf{75.00} (+1.65) \\ \bottomrule
\end{tabular}%
}
\caption{The table presents the pass@1 results of the same model when trained on datasets generated by itself versus those distilled by an external model. In this comparison, the external model used is Meta-Llama-3-70B-Instruct.}
\label{tab:self-gen}
\end{table}

\section{Conclusion}

% 我们提出了用于代码领域的推理增强数据生成流程 SRA-MCTS，能够引导模型自主生成高质量的中间推理路径，从而形成正反馈循环，实现持续改进。该方法在不同模型规模上均优于官方的Instruct版本，并在8B和14B级别上超越了传统的CoT方法。
% We propose a pipeline for the code generation domain, where input questions are processed by SRA-MCTS to generate diverse, high-quality natural language thinking without additional supervision to guide code generation, combine self-generate and self-evaluate to create a positive feedback loop for self-improvement, with experiments showing that self-improved small models outperform those trained with distilled data from a 70B model.
We propose a self-improvement pipeline where SRA-MCTS performs both self-generation and self-evaluation for given questions. The high-quality thinking text generated through this process serves to guide code generation, with both the thinking and the code subsequently used for fine-tuning. Experimental results demonstrate that SRA-MCTS outperforms both CoT and ToT on more complex tasks, achieving an 11-point improvement on the MBPP Complex split.
\appendix

\bibliographystyle{named}
\bibliography{ijcai25}

\begin{thebibliography}{}

\bibitem[\protect\citeauthoryear{Auer \bgroup \em et al.\egroup }{2002}]{DBLP:journals/ml/AuerCF02}
Peter Auer, Nicol{\`{o}} Cesa{-}Bianchi, and Paul Fischer.
\newblock Finite-time analysis of the multiarmed bandit problem.
\newblock {\em Mach. Learn.}, 47(2-3):235--256, 2002.

\bibitem[\protect\citeauthoryear{Austin \bgroup \em et al.\egroup }{2021}]{DBLP:journals/corr/abs-2108-07732}
Jacob Austin, Augustus Odena, Maxwell~I. Nye, Maarten Bosma, Henryk Michalewski, David Dohan, Ellen Jiang, Carrie~J. Cai, Michael Terry, Quoc~V. Le, and Charles Sutton.
\newblock Program synthesis with large language models.
\newblock {\em CoRR}, abs/2108.07732, 2021.

\bibitem[\protect\citeauthoryear{Chaudhary}{2023}]{codealpaca}
Sahil Chaudhary.
\newblock Code alpaca: An instruction-following llama model for code generation.
\newblock \url{https://github.com/sahil280114/codealpaca}, 2023.

\bibitem[\protect\citeauthoryear{Chen \bgroup \em et al.\egroup }{2021}]{DBLP:journals/corr/abs-2107-03374}
Mark Chen, Jerry Tworek, Heewoo Jun, Qiming Yuan, Henrique~Pond{\'{e}} de~Oliveira~Pinto, Jared Kaplan, Harri Edwards, Yuri Burda, Nicholas Joseph, Greg Brockman, Alex Ray, Raul Puri, Gretchen Krueger, Michael Petrov, Heidy Khlaaf, Girish Sastry, Pamela Mishkin, Brooke Chan, Scott Gray, Nick Ryder, Mikhail Pavlov, Alethea Power, Lukasz Kaiser, Mohammad Bavarian, Clemens Winter, Philippe Tillet, Felipe~Petroski Such, Dave Cummings, Matthias Plappert, Fotios Chantzis, Elizabeth Barnes, Ariel Herbert{-}Voss, William~Hebgen Guss, Alex Nichol, Alex Paino, Nikolas Tezak, Jie Tang, Igor Babuschkin, Suchir Balaji, Shantanu Jain, William Saunders, Christopher Hesse, Andrew~N. Carr, Jan Leike, Joshua Achiam, Vedant Misra, Evan Morikawa, Alec Radford, Matthew Knight, Miles Brundage, Mira Murati, Katie Mayer, Peter Welinder, Bob McGrew, Dario Amodei, Sam McCandlish, Ilya Sutskever, and Wojciech Zaremba.
\newblock Evaluating large language models trained on code.
\newblock {\em CoRR}, abs/2107.03374, 2021.

\bibitem[\protect\citeauthoryear{Chen \bgroup \em et al.\egroup }{2024}]{DBLP:conf/cvpr/0003XKISGX24}
Boyuan Chen, Zhuo Xu, Sean Kirmani, Brian Ichter, Dorsa Sadigh, Leonidas~J. Guibas, and Fei Xia.
\newblock Spatialvlm: Endowing vision-language models with spatial reasoning capabilities.
\newblock In {\em {IEEE/CVF} Conference on Computer Vision and Pattern Recognition, {CVPR} 2024, Seattle, WA, USA, June 16-22, 2024}, pages 14455--14465. {IEEE}, 2024.

\bibitem[\protect\citeauthoryear{Coulom}{2006}]{DBLP:conf/cg/Coulom06}
R{\'{e}}mi Coulom.
\newblock Efficient selectivity and backup operators in monte-carlo tree search.
\newblock In H.~Jaap van~den Herik, Paolo Ciancarini, and H.~H. L.~M. Donkers, editors, {\em Computers and Games, 5th International Conference, {CG} 2006, Turin, Italy, May 29-31, 2006. Revised Papers}, volume 4630 of {\em Lecture Notes in Computer Science}, pages 72--83. Springer, 2006.

\bibitem[\protect\citeauthoryear{Dou \bgroup \em et al.\egroup }{2024}]{DBLP:conf/acl/Dou0JZXSHWFXZJZ24}
Shihan Dou, Yan Liu, Haoxiang Jia, Enyu Zhou, Limao Xiong, Junjie Shan, Caishuang Huang, Xiao Wang, Xiaoran Fan, Zhiheng Xi, Yuhao Zhou, Tao Ji, Rui Zheng, Qi~Zhang, Tao Gui, and Xuanjing Huang.
\newblock Stepcoder: Improving code generation with reinforcement learning from compiler feedback.
\newblock In Lun{-}Wei Ku, Andre Martins, and Vivek Srikumar, editors, {\em Proceedings of the 62nd Annual Meeting of the Association for Computational Linguistics (Volume 1: Long Papers), {ACL} 2024, Bangkok, Thailand, August 11-16, 2024}, pages 4571--4585. Association for Computational Linguistics, 2024.

\bibitem[\protect\citeauthoryear{Dubey \bgroup \em et al.\egroup }{2024}]{dubey2024llama}
Abhimanyu Dubey, Abhinav Jauhri, Abhinav Pandey, Abhishek Kadian, Ahmad Al-Dahle, Aiesha Letman, Akhil Mathur, Alan Schelten, Amy Yang, Angela Fan, et~al.
\newblock The llama 3 herd of models.
\newblock {\em arXiv preprint arXiv:2407.21783}, 2024.

\bibitem[\protect\citeauthoryear{Greengerong}{2023}]{greengerong_leetcode}
Greengerong.
\newblock Leetcode dataset, 2023.

\bibitem[\protect\citeauthoryear{Hu \bgroup \em et al.\egroup }{2022}]{DBLP:conf/iclr/HuSWALWWC22}
Edward~J. Hu, Yelong Shen, Phillip Wallis, Zeyuan Allen{-}Zhu, Yuanzhi Li, Shean Wang, Lu~Wang, and Weizhu Chen.
\newblock Lora: Low-rank adaptation of large language models.
\newblock In {\em The Tenth International Conference on Learning Representations, {ICLR} 2022, Virtual Event, April 25-29, 2022}. OpenReview.net, 2022.

\bibitem[\protect\citeauthoryear{Hurst \bgroup \em et al.\egroup }{2024}]{hurst2024gpt}
Aaron Hurst, Adam Lerer, Adam~P Goucher, Adam Perelman, Aditya Ramesh, Aidan Clark, AJ~Ostrow, Akila Welihinda, Alan Hayes, Alec Radford, et~al.
\newblock Gpt-4o system card.
\newblock {\em arXiv preprint arXiv:2410.21276}, 2024.

\bibitem[\protect\citeauthoryear{Kulal \bgroup \em et al.\egroup }{2019}]{DBLP:conf/nips/KulalPC0PAL19}
Sumith Kulal, Panupong Pasupat, Kartik Chandra, Mina Lee, Oded Padon, Alex Aiken, and Percy Liang.
\newblock Spoc: Search-based pseudocode to code.
\newblock In Hanna~M. Wallach, Hugo Larochelle, Alina Beygelzimer, Florence d'Alch{\'{e}}{-}Buc, Emily~B. Fox, and Roman Garnett, editors, {\em Advances in Neural Information Processing Systems 32: Annual Conference on Neural Information Processing Systems 2019, NeurIPS 2019, December 8-14, 2019, Vancouver, BC, Canada}, pages 11883--11894, 2019.

\bibitem[\protect\citeauthoryear{Li \bgroup \em et al.\egroup }{2023}]{DBLP:journals/corr/abs-2305-10679}
Xin{-}Ye Li, Jiang{-}Tian Xue, Zheng Xie, and Ming Li.
\newblock Think outside the code: Brainstorming boosts large language models in code generation.
\newblock {\em CoRR}, abs/2305.10679, 2023.

\bibitem[\protect\citeauthoryear{Li \bgroup \em et al.\egroup }{2024a}]{DBLP:conf/acl/LiZS24}
Haochen Li, Xin Zhou, and Zhiqi Shen.
\newblock Rewriting the code: {A} simple method for large language model augmented code search.
\newblock In Lun{-}Wei Ku, Andre Martins, and Vivek Srikumar, editors, {\em Proceedings of the 62nd Annual Meeting of the Association for Computational Linguistics (Volume 1: Long Papers), {ACL} 2024, Bangkok, Thailand, August 11-16, 2024}, pages 1371--1389. Association for Computational Linguistics, 2024.

\bibitem[\protect\citeauthoryear{Li \bgroup \em et al.\egroup }{2024b}]{DBLP:journals/corr/abs-2409-09584}
Qingyao Li, Wei Xia, Kounianhua Du, Xinyi Dai, Ruiming Tang, Yasheng Wang, Yong Yu, and Weinan Zhang.
\newblock Rethinkmcts: Refining erroneous thoughts in monte carlo tree search for code generation.
\newblock {\em CoRR}, abs/2409.09584, 2024.

\bibitem[\protect\citeauthoryear{Liu \bgroup \em et al.\egroup }{2023}]{DBLP:conf/nips/LiuXW023}
Jiawei Liu, Chunqiu~Steven Xia, Yuyao Wang, and Lingming Zhang.
\newblock Is your code generated by chatgpt really correct? rigorous evaluation of large language models for code generation.
\newblock In Alice Oh, Tristan Naumann, Amir Globerson, Kate Saenko, Moritz Hardt, and Sergey Levine, editors, {\em Advances in Neural Information Processing Systems 36: Annual Conference on Neural Information Processing Systems 2023, NeurIPS 2023, New Orleans, LA, USA, December 10 - 16, 2023}, 2023.

\bibitem[\protect\citeauthoryear{Long \bgroup \em et al.\egroup }{2024}]{DBLP:conf/acl/LongWXZDCW24}
Lin Long, Rui Wang, Ruixuan Xiao, Junbo Zhao, Xiao Ding, Gang Chen, and Haobo Wang.
\newblock On llms-driven synthetic data generation, curation, and evaluation: {A} survey.
\newblock In Lun{-}Wei Ku, Andre Martins, and Vivek Srikumar, editors, {\em Findings of the Association for Computational Linguistics, {ACL} 2024, Bangkok, Thailand and virtual meeting, August 11-16, 2024}, pages 11065--11082. Association for Computational Linguistics, 2024.

\bibitem[\protect\citeauthoryear{Luo \bgroup \em et al.\egroup }{2024}]{DBLP:conf/iclr/LuoX0SGHT0LJ24}
Ziyang Luo, Can Xu, Pu~Zhao, Qingfeng Sun, Xiubo Geng, Wenxiang Hu, Chongyang Tao, Jing Ma, Qingwei Lin, and Daxin Jiang.
\newblock Wizardcoder: Empowering code large language models with evol-instruct.
\newblock In {\em The Twelfth International Conference on Learning Representations, {ICLR} 2024, Vienna, Austria, May 7-11, 2024}. OpenReview.net, 2024.

\bibitem[\protect\citeauthoryear{OpenAI}{}]{openai_learning_to_reason}
OpenAI.
\newblock Learning to reason with large language models.
\newblock \url{https://openai.com/index/learning-to-reason-with-llms/}.
\newblock Accessed: 2024-10-31.

\bibitem[\protect\citeauthoryear{Ouyang \bgroup \em et al.\egroup }{2022}]{DBLP:conf/nips/Ouyang0JAWMZASR22}
Long Ouyang, Jeffrey Wu, Xu~Jiang, Diogo Almeida, Carroll~L. Wainwright, Pamela Mishkin, Chong Zhang, Sandhini Agarwal, Katarina Slama, Alex Ray, John Schulman, Jacob Hilton, Fraser Kelton, Luke Miller, Maddie Simens, Amanda Askell, Peter Welinder, Paul~F. Christiano, Jan Leike, and Ryan Lowe.
\newblock Training language models to follow instructions with human feedback.
\newblock In Sanmi Koyejo, S.~Mohamed, A.~Agarwal, Danielle Belgrave, K.~Cho, and A.~Oh, editors, {\em Advances in Neural Information Processing Systems 35: Annual Conference on Neural Information Processing Systems 2022, NeurIPS 2022, New Orleans, LA, USA, November 28 - December 9, 2022}, 2022.

\bibitem[\protect\citeauthoryear{Team \bgroup \em et al.\egroup }{2024}]{team2024gemma}
Gemma Team, Morgane Riviere, Shreya Pathak, Pier~Giuseppe Sessa, Cassidy Hardin, Surya Bhupatiraju, L{\'e}onard Hussenot, Thomas Mesnard, Bobak Shahriari, Alexandre Ram{\'e}, et~al.
\newblock Gemma 2: Improving open language models at a practical size.
\newblock {\em arXiv preprint arXiv:2408.00118}, 2024.

\bibitem[\protect\citeauthoryear{Team}{2024}]{qwen2.5}
Qwen Team.
\newblock Qwen2.5: A party of foundation models, September 2024.

\bibitem[\protect\citeauthoryear{Wang \bgroup \em et al.\egroup }{2024a}]{DBLP:journals/corr/abs-2409-03733}
Evan Wang, Federico Cassano, Catherine Wu, Yunfeng Bai, Will Song, Vaskar Nath, Ziwen Han, Sean Hendryx, Summer Yue, and Hugh Zhang.
\newblock Planning in natural language improves {LLM} search for code generation.
\newblock {\em CoRR}, abs/2409.03733, 2024.

\bibitem[\protect\citeauthoryear{Wang \bgroup \em et al.\egroup }{2024b}]{DBLP:conf/acl/Wang0DWZDXWZC24}
Yejie Wang, Keqing He, Guanting Dong, Pei Wang, Weihao Zeng, Muxi Diao, Weiran Xu, Jingang Wang, Mengdi Zhang, and Xunliang Cai.
\newblock Dolphcoder: Echo-locating code large language models with diverse and multi-objective instruction tuning.
\newblock In Lun{-}Wei Ku, Andre Martins, and Vivek Srikumar, editors, {\em Proceedings of the 62nd Annual Meeting of the Association for Computational Linguistics (Volume 1: Long Papers), {ACL} 2024, Bangkok, Thailand, August 11-16, 2024}, pages 4706--4721. Association for Computational Linguistics, 2024.

\bibitem[\protect\citeauthoryear{Wei \bgroup \em et al.\egroup }{2022}]{DBLP:conf/nips/Wei0SBIXCLZ22}
Jason Wei, Xuezhi Wang, Dale Schuurmans, Maarten Bosma, Brian Ichter, Fei Xia, Ed~H. Chi, Quoc~V. Le, and Denny Zhou.
\newblock Chain-of-thought prompting elicits reasoning in large language models.
\newblock In Sanmi Koyejo, S.~Mohamed, A.~Agarwal, Danielle Belgrave, K.~Cho, and A.~Oh, editors, {\em Advances in Neural Information Processing Systems 35: Annual Conference on Neural Information Processing Systems 2022, NeurIPS 2022, New Orleans, LA, USA, November 28 - December 9, 2022}, 2022.

\bibitem[\protect\citeauthoryear{Wei \bgroup \em et al.\egroup }{2023}]{DBLP:journals/corr/abs-2312-02120}
Yuxiang Wei, Zhe Wang, Jiawei Liu, Yifeng Ding, and Lingming Zhang.
\newblock Magicoder: Source code is all you need.
\newblock {\em CoRR}, abs/2312.02120, 2023.

\bibitem[\protect\citeauthoryear{Yao \bgroup \em et al.\egroup }{2023}]{DBLP:conf/nips/YaoYZS00N23}
Shunyu Yao, Dian Yu, Jeffrey Zhao, Izhak Shafran, Tom Griffiths, Yuan Cao, and Karthik Narasimhan.
\newblock Tree of thoughts: Deliberate problem solving with large language models.
\newblock In Alice Oh, Tristan Naumann, Amir Globerson, Kate Saenko, Moritz Hardt, and Sergey Levine, editors, {\em Advances in Neural Information Processing Systems 36: Annual Conference on Neural Information Processing Systems 2023, NeurIPS 2023, New Orleans, LA, USA, December 10 - 16, 2023}, 2023.

\bibitem[\protect\citeauthoryear{Yu \bgroup \em et al.\egroup }{2024}]{DBLP:conf/acl/YuZSHXZHY24}
Zhaojian Yu, Xin Zhang, Ning Shang, Yangyu Huang, Can Xu, Yishujie Zhao, Wenxiang Hu, and Qiufeng Yin.
\newblock Wavecoder: Widespread and versatile enhancement for code large language models by instruction tuning.
\newblock In Lun{-}Wei Ku, Andre Martins, and Vivek Srikumar, editors, {\em Proceedings of the 62nd Annual Meeting of the Association for Computational Linguistics (Volume 1: Long Papers), {ACL} 2024, Bangkok, Thailand, August 11-16, 2024}, pages 5140--5153. Association for Computational Linguistics, 2024.

\bibitem[\protect\citeauthoryear{Zhang \bgroup \em et al.\egroup }{2024a}]{DBLP:journals/corr/abs-2406-03816}
Dan Zhang, Sining Zhoubian, Yisong Yue, Yuxiao Dong, and Jie Tang.
\newblock Rest-mcts*: {LLM} self-training via process reward guided tree search.
\newblock {\em CoRR}, abs/2406.03816, 2024.

\bibitem[\protect\citeauthoryear{Zhang \bgroup \em et al.\egroup }{2024b}]{DBLP:journals/corr/abs-2410-10762}
Jiayi Zhang, Jinyu Xiang, Zhaoyang Yu, Fengwei Teng, Xionghui Chen, Jiaqi Chen, Mingchen Zhuge, Xin Cheng, Sirui Hong, Jinlin Wang, Bingnan Zheng, Bang Liu, Yuyu Luo, and Chenglin Wu.
\newblock Aflow: Automating agentic workflow generation.
\newblock {\em CoRR}, abs/2410.10762, 2024.

\bibitem[\protect\citeauthoryear{Zheng \bgroup \em et al.\egroup }{2024}]{zheng2024llamafactory}
Yaowei Zheng, Richong Zhang, Junhao Zhang, Yanhan Ye, Zheyan Luo, Zhangchi Feng, and Yongqiang Ma.
\newblock Llamafactory: Unified efficient fine-tuning of 100+ language models.
\newblock In {\em Proceedings of the 62nd Annual Meeting of the Association for Computational Linguistics (Volume 3: System Demonstrations)}, Bangkok, Thailand, 2024. Association for Computational Linguistics.

\end{thebibliography}

\end{document}